\documentclass[runningheads]{style-eccv/llncs}

\usepackage{style-eccv/eccv}

\usepackage{style-eccv/eccvabbrv}
\usepackage{graphicx}
\usepackage{booktabs}
\usepackage[accsupp]{axessibility}%

\usepackage{hyperref}
\usepackage{orcidlink}

\usepackage{ulem}

\usepackage{multirow}
\usepackage{makecell}
\renewcommand{\arraystretch}{1.1}

\newcommand{\mr}[2]{\multirow{#1}{*}{#2}}

\usepackage{xparse}
\usepackage{pifont}

\usepackage{textcomp}%
\usepackage{scalerel}%

\usepackage{tabularx,colortbl}
\usepackage{placeins}

\usepackage{enumitem}

\usepackage[ruled,linesnumbered,vlined]{algorithm2e}
\newcommand{\method}{UniTemp\xspace}

\newcommand{\cmark}{\ding{51}}%
\newcommand{\xmark}{\ding{55}}%

\newlength\savewidth
\newlength\thinwidth

\definecolor{Gray}{gray}{0.93}
\definecolor{ForestGreen}{rgb}{0.13, 0.75, 0.13}

\usepackage{xspace}%
\makeatletter
\DeclareRobustCommand\onedot{\futurelet\@let@token\@onedot}
\def\@onedot{\ifx\@let@token.\else.\null\fi\xspace}
\def\eg{\textit{e.g}\onedot} 
\def\ie{\textit{i.e}\onedot}

\makeatother

\usepackage{amsmath,amsfonts,bm}

\def\1{\bm{1}}

\DeclareMathAlphabet{\mathsfit}{\encodingdefault}{\sfdefault}{m}{sl}
\SetMathAlphabet{\mathsfit}{bold}{\encodingdefault}{\sfdefault}{bx}{n}

\newcommand{\E}{\mathbb{E}}

\newcommand{\KL}{D_{\mathrm{KL}}}

\AtBeginDocument{%
  \setlength{\abovedisplayskip}{3pt plus 1pt minus 1pt}%
  \setlength{\belowdisplayskip}{3pt plus 1pt minus 1pt}%
  \setlength{\abovedisplayshortskip}{2pt plus 1pt minus 1pt}%
  \setlength{\belowdisplayshortskip}{2pt plus 1pt minus 1pt}%
  \setlength{\jot}{2pt}%
}

\title{UniTemp: Unlocking Video Generation in Any Temporal Order via Bidirectional Distillation}

\author{
Lin Zhang\inst{1}$^{*}$ \and
Sicheng Mo\inst{3} \and
Zefan Cai\inst{1} \and
Jinhong Lin\inst{1} \and
Zihao Lin\inst{4} \and
Jiuxiang Gu\inst{2} \and
Krishna Kumar Singh\inst{2} \and
Yuheng Li\inst{2}$^{\dagger}$ \and
Yin Li\inst{1}$^{\dagger}$
}
\authorrunning{L.~Zhang et al.}
\titlerunning{UniTemp}

\institute{
University of Wisconsin Madison \and
Adobe Research \and
University of California Los Angeles \and
University of California Davis\\[0.3em]
\url{https://lzhangbj.github.io/projects/unitemp/}
}

\begin{document}

\begingroup
\renewcommand{\addcontentsline}[3]{}
\renewcommand{\addtocontents}[2]{}

\maketitle
\renewcommand{\thefootnote}{}
\footnotetext{$^*$Work partially done during internship at Adobe Research.}
\footnotetext{$^\dagger$Equal advising.}
\renewcommand{\thefootnote}{\arabic{footnote}}

\begin{figure}[t]
\centering
\includegraphics[width=0.95\linewidth]{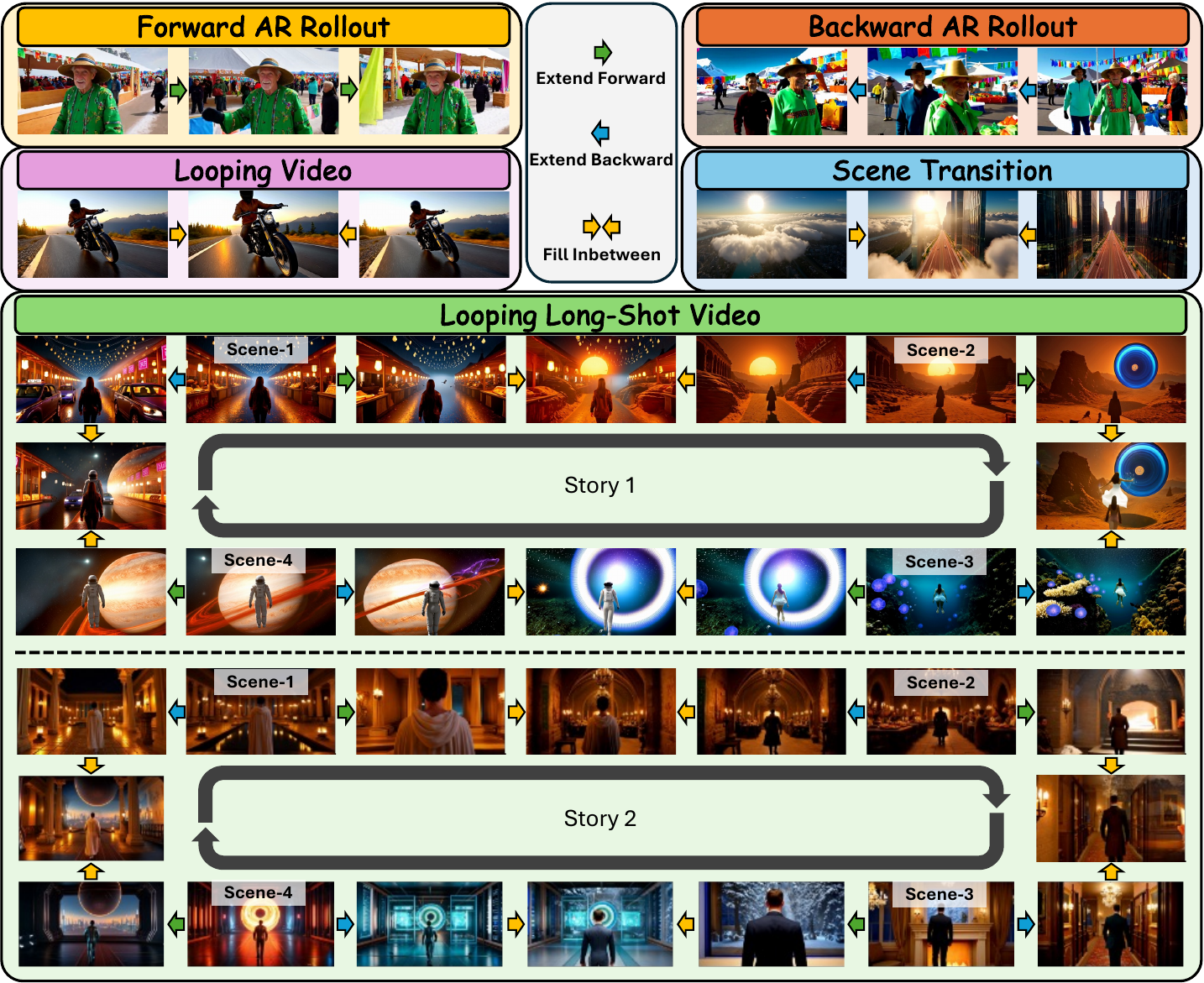}\vspace{-0.5em}
\caption{We present \textbf{\method}, a unified distillation framework that delivers a single model capable of flexibly generating video conditioned on past context, future context, or both, and supporting a wide range of generation tasks.}\vspace{-2.5em}
\label{fig:teaser}
\end{figure}

\begin{abstract}

Autoregressive video diffusion models have emerged as a promising approach for long video generation, achieving strong performance in streaming settings.
However, existing methods are restricted to forward temporal generation, whereas practical video creation often requires flexible generation order, e.g., conditioning on future context to extend backward, or on both past and future context for inbetween generation.
We bridge this gap by training an autoregressive model that supports generation in arbitrary temporal directions.
A key technical challenge arises from the Causal 3D VAE widely used in video diffusion models, which encodes latents strictly conditioned on past context.
While suited for forward generation, this causal structure causes inter-block discontinuities when generation proceeds backward.
To address this, we introduce blockwise \textit{anchor latents}, a set of auxiliary latents that restore the missing past context at block boundaries during backward generation.
Built on this design, we propose \method, a bidirectional distillation framework that trains a single autoregressive student model for any-direction video generation.
At inference time, \method conditions on arbitrary past and/or future frames, improving controllability for both bidirectional and inbetween generation.
Experiments show that \method maintains competitive performance on short and long video generation compared to forward-only methods, while enabling diverse workflows such as bidirectional video extension, inbetween generation, looping video generation, scene transition, and visual story generation.


\end{abstract}

\section{Introduction}\label{sec:intro}

Diffusion transformers~\cite{dit,diffusion,transformer} have dominated the video generation landscape, achieving high-quality synthesis across diverse visual domains~\cite{wan2024,sora2024,moviegen2024,cogvideox2025,cosmos2025,hunyuan2025,opensora2024}. Despite their impressive performance, these models require multiple denoising steps with full-sequence attention at inference, making them computationally expensive and difficult to scale to long video generation.
Recently, asymmetric distillation~\cite{dmd,dmdv2} has been introduced to address these limitations by distilling a large multi-step full-attention teacher model into a few-step autoregressive student. By generating videos block by block and caching key-value (KV) pairs, such models enable efficient streaming video generation~\cite{causvid2024}.

Existing autoregressive video generators support only \textit{forward}~(causal) video generation, where each block is conditioned exclusively on generated past content.
Real-world video creation workflows, however, often require flexible temporal ordering and iterative editing of generated content.
For example, a user may wish to generate a prologue for an existing clip (backward extension), fill a temporal gap between two segments (inbetween generation), or generate key frames first then connect them into a visual story.
While solutions have been developed~\cite{svd2024,streamingt2v2025,gi2024,mcvd2022,vace2025,tokenflow2024,tanveer2025multicoin}, they each address a specific temporal conditioning mode.
An open question remains: \textit{can we develop a unified autoregressive model capable of flexibly generating video conditioned on past context, future context, or both, thereby supporting a wide range of generation tasks at real-time speed?}

In this work, we aim to enable autoregressive video generation in any temporal direction.
A natural starting point is to extend existing forward-only asymmetric distillation~\cite{selfforcing} to also support backward generation.
However, we find that such a naive approach leads to noticeable inter-block discontinuities and boundary flickering in the generated content, particularly in scenes with high motion dynamics.
We trace the root cause of these artifacts to the \textit{Causal 3D VAE}~\cite{magvitv22025}, a component widely adopted in modern video diffusion models~\cite{wan2024,cogvideox2025,cosmos2025,hunyuan2025,opensora2024} for efficient spatio-temporal encoding.
Causal 3D VAE encodes video frames into latents that are strictly conditioned on its preceding context.
While this causal structure aligns naturally with forward generation, it creates a fundamental mismatch for backward generation: 
when generating backward, the past context that the decoder expects is unavailable, causing visible discontinuities at block boundaries.
Retraining a non-causal VAE is impractical, as it would forfeit compatibility with existing pre-trained teacher models.

To address this issue, we introduce blockwise \textbf{anchor latents}, a mechanism that restores the missing past context at block boundaries during backward generation.
The key intuition is to jointly denoise a small set of auxiliary latents standing for preceding video tokens alongside the target block under full-sequence attention, allowing the generated frames to attend to a proxy of prior local context even when generating in reverse order.
These anchor latents act as a bridge between consecutive blocks, providing smooth, artifact-free transitions without requiring any modification to the pre-trained VAE or teacher model.

Built on our design, we present \textbf{\method}, a unified distillation framework that trains a single autoregressive student model, supervised by a frozen teacher diffusion model, to support generation in any direction.
\method adopts an efficient training procedure via shared models between forward and backward generation tasks.
During inference, \method flexibly conditions on past context, future context, or both, enabling any-direction video generation and unlocking a diverse set of applications all in one model (see Fig.~\ref{fig:teaser}).

\noindent \textbf{Our contributions} are summarized as follows.
\begin{enumerate}[nosep]
    \item We introduce \method, a unified distillation framework that enables autoregressive video generation in any temporal direction. The framework maintains training efficiency while delivering a single model for flexible test-time conditioning with fast inference.
    \item We show that directly training on backward generation leads to severe inter-block flickering. We trace this artifact to the causal structure of the Causal 3D VAE, then introduce blockwise anchor latents that restore the missing past context at block boundaries, substantially reducing visual artifacts.
    \item We demonstrate competitive performance on short, long, and inbetween video generation compared to forward-only methods, while unlocking new capabilities such as bidirectional video extension, scene transition, looping video generation, and visual story generation.
\end{enumerate}

\section{Related Work}\label{sec:related-work}

\noindent\textbf{Temporal dependency in video generation.}
Video generation tasks differ in their temporal conditioning: text-to-video~\cite{wan2024,sora2024,moviegen2024} generates from scratch, video extension~\cite{svd2024,streamingt2v2025} conditions on the past, and inbetween generation~\cite{gi2024,explorative2024,mcvd2022,film2022} conditions on both endpoints.
Supporting all these modes in a single model remains an open challenge. Prior methods for flexible conditioning adopt mask-guided inpainting with full-sequence attention.
MCVD~\cite{mcvd2022} concatenates past and future frames with noisy targets along the channel dimension, randomly masking past or future frames to handle both extension and interpolation.
RaMViD~\cite{ramvid2022} randomly masks subsets of frames during training, applying diffusion only to the masked frames to enable conditioning on arbitrary positions.
VACE~\cite{vace2025} unified a broad range of generation and editing tasks through an input video mask that distinguishes guided regions from generated ones.
However, these approaches require full-sequence denoising with sophisticated input formats (\eg, masks) and redundant feature extraction, and thus are slow at inference. They also pose training difficulties because they require massive real video data for pretraining or fine-tuning.
Our method instead unifies these tasks within a single distillation framework. During training, this distillation approach remains efficient and does not require real video data, unlike previous pretraining- or fine-tuning-based approaches. During inference, we adopt temporal position indices in RoPE~\cite{rope} to specify conditioning positions, and use KV caches to avoid redundant computation over previously generated blocks, thereby achieving much faster inference.

\medskip
\noindent\textbf{VAE in latent video diffusion models.}
Generating videos directly in pixel space is computationally prohibitive.
Latent diffusion models~\cite{ldm2022} address this by compressing videos into a compact latent space via a variational autoencoder (VAE) and training a diffusion model in that space.
State-of-the-art models~\cite{wan2024,sora2024,cogvideox2025,hunyuan2025} adopt diffusion transformers (DiT)~\cite{dit} as the backbone, operating on spatio-temporal latent tokens with self-attention and text conditioning via cross-attention.
A critical design choice is the VAE architecture.
Causal 3D VAEs~\cite{magvitv22025} have become the dominant choice for their advantages in
(1)~unified image and video tokenization,
(2)~improved generation quality, and
(3)~reduced memory usage,
and are widely adopted by state-of-the-art video diffusion models including CogVideoX~\cite{cogvideox2025}, Cosmos~\cite{cosmos2025}, HunyuanVideo~\cite{hunyuan2025}, and Wan~\cite{wan2024}.
However, a causal 3D VAE introduces a temporal bias: each latent's encoding and decoding depend only on its past context.
This bias, which has not been well studied in prior work, becomes a critical obstacle when generation proceeds backward, as we will show in \cref{sec:method:anchor}.
In this work, we address this issue with blockwise anchor latents, a sequence of preceding video latents that participate in the denoising of each block, effectively stabilizing reverse block generation and largely reducing artifacts.

\medskip
\noindent\textbf{Autoregressive video generation via distillation.}
Pretraining video generation models on massive long-video datasets is challenging.
Asymmetric distillation~\cite{causvid2024} was proposed as a post-training technique that distills a multi-step full-attention teacher trained on short horizons into a few-step autoregressive student for long video generation. The student generates videos block by block with KV caching, enabling efficient streaming synthesis for long videos.
Self-Forcing~\cite{selfforcing} further bridges the train-test distribution gap by conditioning on the student's own rollouts rather than ground truth videos during training, largely reducing quality degradation over long horizons.
Subsequent work has stabilized long generation through training-based~\cite{longlive2025,rollingforcing2025,rewardforcing,selfforcing++} and training-free~\cite{infinityrope2025,infiniteforcing2025,deepforcing} techniques.
These methods, however, support only forward generation.
Our \method builds on the Self-Forcing framework for efficient training but departs from prior work by training a single student for any-direction generation.

\vspace{-0.1em}
\section{Preliminaries}\label{sec:preliminary}

\subsection{Latent Video Diffusion Models and Causal 3D VAE}\label{sec:preliminary:causal-vae}
Latent video diffusion models~\cite{ldm2022}~(LVDMs) encode frames $\{x_i\}$ into VAE latents $\{z_i\}$, then use a denoiser~(\eg, U-Net~\cite{unet} or DiT~\cite{dit}) to generate latents from noise.
At each noise level $t$, the denoiser maps noisy latents $\{z_i^t\}$ to cleaner latents $\{z_i^{t-1}\}$ conditioned on text.

Most recent LVDMs adopt a Causal 3D VAE~\cite{magvitv22025,wan2024,cogvideox2025,hunyuan2025,cosmos2025,opensora2024}, which uses an encoder $\mathcal{E}$ and a decoder $\mathcal{D}$ to encode and decode each latent $z_i$ using its past context $z_{\leq i}$, as shown in \cref{fig:method}.
With a temporal compression factor of $4$, $z_0$ is a special image latent encoding the initial frame $x_0$, and $z_i$~($i>0$) encodes every four consecutive frames $(x_{4i},\ldots,x_{4i+3})$.
This supports streaming and joint image-video tokenization~\cite{magvitv22025}, but creates a forward temporal bias~(\cref{sec:method:anchor}).

\subsection{Self-Forcing}\label{sec:preliminary:autoregressive-distillation}

Self-Forcing~\cite{selfforcing} distills a slow teacher, such as Wan2.1~\cite{wan2024}, into a few-step autoregressive student $G^\theta$ through two training stages.
Stage 1 initializes $G^\theta$ with $L$ teacher-generated ODE trajectories $\{z_l^t\}_{l=1}^L$, where $t$ covers the student's denoising steps $\{t_i\}_{i=1}^N$, by regressing each noisy latent to its clean target:
\begin{equation}
L_{\text{init}} = \E_{z^0,i} \left\| G^\theta(z^{t_i}, t_i) - z^0 \right\|_2^2
\end{equation}

where $z^0$ denotes the clean latent and $z_{t_i}$ is the noisy latent at denoising step $t_i$ along the teacher's ODE trajectory.
Stage 2 trains $G^\theta$ in the same blockwise autoregressive rollout used at inference, where each block is denoised conditioned on previously generated blocks rather than ground truth.
For block size $B=3$, this follows $(z_0,z_1,z_2) \rightarrow (z_3,z_4,z_5) \rightarrow \cdots \rightarrow (z_{18},z_{19},z_{20})$.
Distribution matching distillation (DMD)~\cite{dmd,dmdv2} matches the student's output distribution to the teacher's by updating $G^\theta$ with:
\begin{equation}\label{eq:dmd}
\nabla_\theta \KL = \E_{\substack{z \sim \mathcal{N}(0;I), x = G^\theta(z)}} \left( s_\text{real}(x) - s_\text{fake}(x) \right) \frac{\partial G^\theta}{\partial \theta}
\end{equation}

where $s_\text{real}$ is computed from the frozen teacher and $s_\text{fake}$ from a fake denoiser $\mu_{\text{fake}}^\phi$ trained to approximate the student's distribution:
\begin{equation}
L_{\text{denoise}}^{\phi} = \left\| \mu_{\text{fake}}^\phi(z^{t_i}, t_i) - z^0 \right\|_2^2
\end{equation}

As in GAN training~\cite{gan}, $\mu_{\text{fake}}^\phi$ is updated with $L_{\text{denoise}}^{\phi}$, while $G^\theta$ is updated with \cref{eq:dmd} to match the teacher, as shown in \cref{fig:method}.

\section{Methods}\label{sec:method}

Existing autoregressive distillation supports only forward generation, which limits practical video creation workflows.
We aim to distill a student model that supports generation in any temporal order at test time.
In \cref{sec:method:anchor}, we first describe a critical issue: inter-block flickering in backward generation, which results from the latent causality induced by the causal 3D VAE.
We then propose blockwise anchor latents as our solution to effectively stabilize backward generation.
Finally, in \cref{sec:method:unified-training}, we discuss our unified distillation framework, as well as its versatility in any-order generation at test time.
By default, we use a block size of $B=3$ in this section.

\subsection{Blockwise Anchor Latents for Backward Generation}\label{sec:method:anchor}

A unified generation framework necessitates backward generation capability.
An intuitive approach is to extend Self-Forcing~\cite{selfforcing} for backward generation by reversing the student's block generation order, i.e.,
  $(z_0, z_1, z_2) \leftarrow \cdots \leftarrow (z_{15}, z_{16}, z_{17}) \leftarrow (z_{18}, z_{19}, z_{20})$, with each block generated conditioned on its future blocks.
In practice, however, we observe severe flickering in backward-generated videos, as shown in \cref{fig:inter-block-flickering}.
This flickering appears periodically at block boundaries, often as ghosting/flashing artifacts covering the whole image, and is more obvious in high-motion local regions.

\medskip
\noindent\textbf{Evaluating periodic flickering.}
To better understand this issue, we need a metric to evaluate these flickering artifacts.
Temporal flickering in VBench~\cite{vbench} serves as a good starting point, but with two drawbacks:
  (1) it averages frame-frame pixel differences over the whole video rather than in local temporal context, and thus fails to identify \emph{when} the flickering happens,
  (2) it is highly influenced by video dynamics, where high-dynamic scenes are more likely to have lower flickering scores.
We thus define Flickering Ratio on top of temporal flickering to address these two problems.
For two neighboring latents $z_i$ and $z_{i+1}$ corresponding to frames $(x_{4i},x_{4i+1},x_{4i+2},x_{4i+3})$ and $(x_{4i+4},x_{4i+5},x_{4i+6},x_{4i+7})$, respectively, we compute the flickering ratio (FR) as:
\begin{equation}\label{eq:flickering-ratio}
  \text{FR}(z_i, z_{i+1}) = \frac{\sum_{h,w,c} |x_{4i+4}(h,w,c) - x_{4i+3}(h,w,c)|}{\sum_{h,w,c} |x_{4i+3}(h,w,c) - x_{4i+2}(h,w,c)|},
\end{equation}

The numerator measures the pixel L1 distance between decoded frames at latent boundary $x_{4i+3} \rightarrow x_{4i+4}$.
The denominator measures the pixel L1 difference between two frames decoded by the same latent, which serves as a good indicator of dynamics in the local clip.
By dividing, we measure cross-latent flickering relative to the local dynamics, which is more robust than simple pixel differences.

We sample a set of dynamic prompts to measure FR.
For each video, we split FR into the following two groups and average the value within each group:
  (1) inter-block FR, where $z_i$ and $z_{i+1}$ belong to different generated blocks, and
  (2) intra-block FR, where $z_i$ and $z_{i+1}$ belong to the same generated block.
Specifically, since intra-block latents can mutually attend to each other with full-sequence self-attention at every layer, they are expected to demonstrate higher temporal smoothness, thus lower FR. Inter-block latents can only attend in one direction through cached keys and values, and therefore should demonstrate higher FR.
We validate this empirically in \cref{tab:inter-block-flickering}, where both forward and backward generation show inter-block FRs larger than 1 ($1.15/1.42$), while intra-block FRs remain at smaller values ($0.93/0.89$).

\begin{figure}[t]
  \vspace{0pt}
  \centering
  \includegraphics[width=\linewidth]{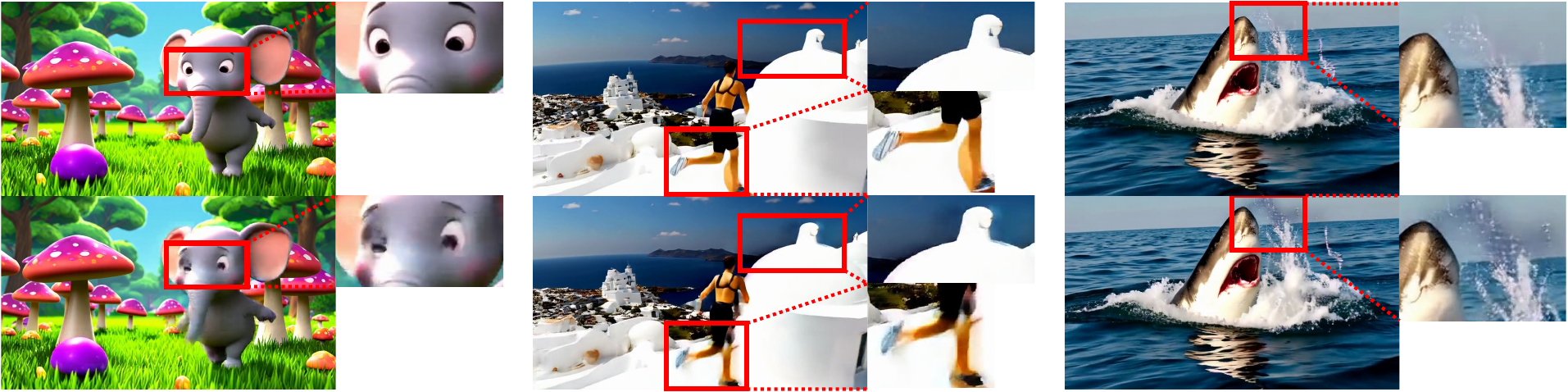}
  \captionof{figure}{Visualization of \textbf{inter-block flickering in backward generation}. Without anchor latents, visible discontinuities appear at block boundaries.}
  \label{fig:inter-block-flickering}
\end{figure}

\medskip
\noindent\textbf{Flickering due to causal 3D VAE.}
The abnormally high inter-block FR in backward generation ($1.42$) aligns with our aforementioned empirical observation of cross-block flickering.
We attribute this to the causal dependency in the encoded latents $\{z_i\}$, which is introduced by a Causal 3D VAE.
A Causal 3D VAE imposes a strong dependency of each generated latent $z_i$ on its past context $z_{<i}$ but not future context $z_{>i}$, as shown in Fig.\ \ref{fig:method} left.
In the forward process, generation of each latent $z_i$ naturally follows this causality since it can always attend to past latents $z_{<i}$.
However, in backward generation, the latent $z_i$ near the left boundary of a block $(z_i, z_{i+1}, z_{i+2})$ can only attend to its future context $z_{>i}$, without past context.
In contrast, full-sequence attention is applied within the block so that $z_{i+1}$ and $z_{i+2}$ have local past context $(z_i)$ and $(z_i, z_{i+1})$, respectively.
Intra-block FR in backward generation can therefore maintain a low value, as in forward generation.

\medskip
\noindent\textbf{Blockwise anchor latents.}
Motivated by our analysis, we seek to solve this inter-block flickering issue via an approximation of the missing past context.
Specifically, in every backward-generated block, we expand its block size from $B$ to $(P + B)$, where $P$ is the number of anchor latents.
Each block is now $(z_{i-P}, ... , z_{i-1}, z_i, z_{i+1}, z_{i+2})$ instead of $(z_i, z_{i+1}, z_{i+2})$.
This expanded block is jointly denoised with full-sequence self-attention within the block, while attending to stored KV cache entries of future latents $z_{>i+2}$ to receive future conditioning.
Note that while these anchor latents $(z_{i-P}, ..., z_{i-1})$ can fill in the missing past context for $z_i$, they themselves (especially $z_{i-P}$) now stay near the left boundary of the block, lacking their own past context.
Therefore, after denoising, we discard the denoised anchor latents and do not include them as outputs. This idea is illustrated in Fig.\ \ref{fig:method} right.

We empirically verify that these anchor latents effectively reduce inter-block flickering, as shown in \cref{tab:inter-block-flickering}. We defer the detailed discussion of our validation to \cref{sec:experiments:ablation}.

\subsection{Unified Training with Bidirectional Distillation}\label{sec:method:unified-training}

\begin{figure}[t]
  \centering
  \includegraphics[width=\linewidth]{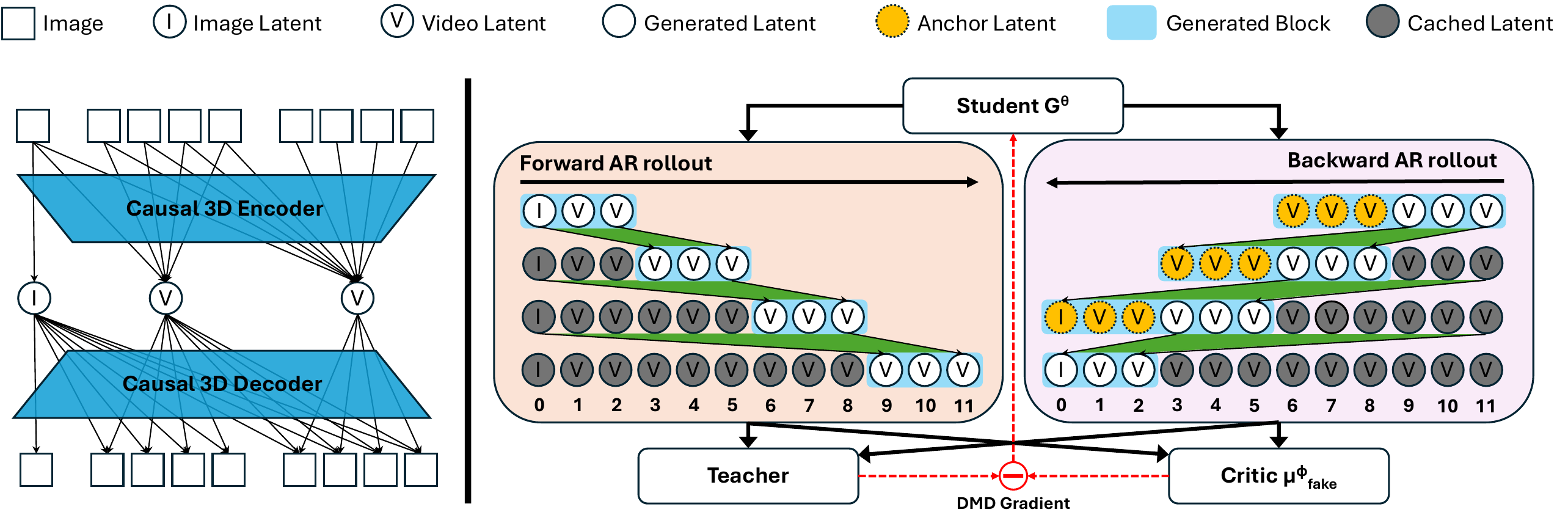}
  \caption{
    \textbf{Left}: Causal design of the frozen 3D VAE. It encodes video into spatial-temporal latents (V) with a leading image latent (I). Each latent is dependent on its past context.
    \textbf{Right}: Overview of \textbf{\method}. We distill a teacher model into a unified autoregressive student $G^\theta$ trained on its self-rollout in both forward and backward directions.
    In backward generation, we introduce \textbf{blockwise anchor latents (dashed circles)} to reduce inter-block flickering.
    The anchor latents only serve to stabilize generated content by providing approximate missing past context.
    After being denoised with the current block, they are discarded and not included as outputs.
  }
  \label{fig:method}
\end{figure}

We now integrate blockwise anchor latents into the following unified framework for any-order temporal generation.

\medskip
\noindent\textbf{Bidirectional distillation.}
During training, we adopt Self-Forcing~\cite{selfforcing} as the training algorithm and train on both forward and backward tasks, as shown in \cref{fig:method}.
The forward generation pipeline faithfully follows the original Self-Forcing with blockwise ($B=3$) autoregressive generation.
In backward generation, we introduce anchor latents to expand the block size from $B$ to $(P+B)$, while the DMD~\cite{dmd,dmdv2} loss is applied to the student's output without anchor latents.

\medskip
\noindent\textbf{Unified models.}
Since training includes two directional targets, one option is to use separate models for forward and backward generation, i.e.,
(1) separate generators. We can use two generators $G^\theta_{\rightarrow}$ and $G^\theta_{\leftarrow}$ to generate forward and backward videos, respectively,
(2) separate fake critics. We can use two fake critics $\mu_{\text{fake},\rightarrow}^{\phi}$ and $\mu_{\text{fake},\leftarrow}^{\phi}$ to approximate the distributions of the forward- and backward-generated videos, respectively.
However, separating models as in previous methods~\cite{gi2024} incurs potentially higher memory cost during training and inference. Separate generators also lose the ability to support flexible temporal conditioning (\eg, inbetween generation) at test time.
Therefore, we adopt a unified model design, sharing a single student model $G^\theta$ and a single fake critic model $\mu_{\text{fake}}^\phi$ across both generation tasks.
Such a shared design can also reduce the performance gap between forward and backward generation, as shown in \cref{sec:experiments:ablation}.

\subsection{Versatile Test-Time Functions}\label{sec:method:versatile-test}

By training a shared generator in both directions, \method allows generating each block conditioned on any combination of past and future KV caches, thereby unlocking
versatile test-time functions as follows:

\medskip
\noindent\textbf{Both-end-guided sink latents.}
Sink latents~\cite{xiaoefficient} have proven to significantly improve inference performance and efficiency over long horizons.
While previous methods~\cite{longlive2025,rollingforcing2025,infinityrope2025,deepforcing,rewardforcing,selfforcing++,infiniteforcing2025} can only place sink latents at past positions to stabilize long video generation,
we can now shift these latents into the future by reapplying their RoPE~\cite{rope,infinityrope2025,infiniteforcing2025} temporal indices, simulating future sink latents for forward generation.
Similarly, in backward generation, we can reposition earlier generated future latents into the past.
This yields both-end-guided sink latents, further reducing quality drift and content variation in long video generation.

\medskip
\noindent\textbf{Inbetween generation.}
Surprisingly, we observe that, although the block size is set to $B$ (or $P+B$ with anchor latents) during training, the test-time block size
can be flexibly set to any value as long as the maximum attention horizon is less than that in training, \ie 21.
This enables inbetween generation~\cite{gi2024,explorative2024,wan2024,film2022,mcvd2022,ldmvfi2024}, which infills the content of $K$ latents between a head block and a tail block.
Specifically, given a head block of $B$ latents and a tail block of $B$ latents, we can first obtain their KV caches by feeding them to the denoiser with clean context noise and a frame gap of $K$ in their RoPE temporal indices.
Then, we can produce the intermediate $K$ latents from noise by applying full-sequence self-attention within the $K$ latents and attending to the head and tail KV caches.

We evaluate and show results in~\cref{sec:exp:results}.
Furthermore, we demonstrate diverse applications in~\cref{sec:experiments:application}, showing flexible temporal generation including looping video generation, scene transition, and visual story generation.

\section{Experiments and Results}\label{sec:experiments}

\noindent\textbf{Training details.}
We train \method following the training procedure of Self-Forcing~\cite{selfforcing}.
Specifically, we use Wan2.1 T2V 14B~\cite{wan2024} as the pretrained teacher model. The student is a 1.3B model initialized from Wan2.1 T2V 1.3B with 4 denoising steps $(1000.0, 937.5, 833.3, 625.0)$.
In the first stage, we use the teacher to generate 16K ODE pairs, and train the student model for 6K iterations with a batch size of 128.
Both forward and backward attention masks are applied in each training iteration.
In the second stage, we train the student for 600 iterations with a batch size of 64 on the rewritten VidProm~\cite{vidprom} prompts.
We apply exponential moving average (EMA) to the student model with a decay rate of 0.999 starting from iteration 200.
We choose a block size of $B=3$ and an anchor latent size of $P=3$ for backward generation, unless otherwise specified.

\medskip
\noindent\textbf{Evaluation protocol.}
We primarily evaluate using metrics in VBench~\cite{vbench}. While our trained model supports a diverse set of applications, we focus on the following dimensions:
(1) Flickering ratio.
This is the proposed metric described in \cref{sec:method:anchor}.
Since flickering is more obvious in high-dynamic video, we use an LLM to generate 128 high-dynamic video prompts to evaluate the flickering ratio.
(2) Single-direction short video generation.
Following conventional evaluation protocol in previous works~\cite{selfforcing,longlive2025},
we generate 5s videos using 946 official VBench prompts rewritten by Qwen2.5-7B-Instruct~\cite{qwen25instruct}.
Metrics are computed in the official VBench codebase.
(3) Single-direction long video generation.
Following previous approaches~\cite{rollingforcing2025,selfforcing++,infinityrope2025},
we randomly sample 128 prompts from MovieGenBench~\cite{moviegenbench} to generate 100s long videos for each prompt. We then use metrics in VBench to evaluate video quality at 10s, 30s, and 100s.
We compare to a set of strong long video generation baseline approaches based on Self-Forcing~\cite{selfforcing}, including LongLive~\cite{longlive2025}, Rolling-Forcing~\cite{rollingforcing2025}, and Infinity-Rope~\cite{infinityrope2025}.
(4) Inbetween video generation.
We use Wan2.1 T2V 14B to generate 5s ground-truth videos using the same 128 prompts from MovieGenBench.
We keep the head block (9 frames) and tail block (12 frames) as conditions, and fill the inbetween region.
In addition to VBench metrics, we additionally measure
\emph{head / tail frame flickering}, the pixel consistency between generated and ground-truth condition frames, and
\emph{FID / FVD} between generated and ground-truth videos.
We compare to Wan FLF2V 14B~\cite{wan2024} and Generative Inbetweening (GI)~\cite{gi2024}.

\subsection{Results and Discussion}\label{sec:exp:results}

\begin{table*}[t]
\begin{minipage}[t]{0.47\textwidth}
\centering
\captionof{table}{\textbf{Ablation study on anchor latents} for backward generation.}
\label{tab:inter-block-flickering}
\resizebox{\linewidth}{!}{%
\setlength{\tabcolsep}{4pt}
\begin{tabular}{c c c c c c}
\toprule
\multirow{2}{*}{Direction} & \multirow{2}{*}{\shortstack{Num Anchor $P$}} & \multicolumn{2}{c}{Anchor as Output} & \multirow{2}{*}{\shortstack{Inter-block\\FR $\downarrow$}} & \multirow{2}{*}{\shortstack{Intra-block\\FR $\downarrow$}} \\
\cmidrule(lr){3-4}
 & & Train & Inference & & \\
\midrule
$\rightarrow$ & 0 & \xmark & \xmark & 1.15 & 0.93 \\
$\leftarrow$  & 0 & \xmark & \xmark & 1.42 & \textbf{0.89} \\
$\leftarrow$  & 1 & \xmark & \xmark & 1.26 & 0.93 \\
$\leftarrow$  & 2 & \xmark & \xmark & 1.23 & 0.93 \\
$\leftarrow$  & 3 & \xmark & \xmark & \textbf{1.07} & 0.97 \\
$\leftarrow$  & 3 & \xmark & \cmark & 1.46 & 0.99 \\
$\leftarrow$  & 3 & \cmark & \xmark & 1.08 & 0.97 \\
$\leftarrow$  & 3 & \cmark & \cmark & 1.48 & 0.98 \\
\bottomrule
\end{tabular}%
}
\end{minipage}%
\hfill
\begin{minipage}[t]{0.45\textwidth}
\centering
\captionof{table}{Ablation study on training with shared models. Results evaluated with VBench~\cite{vbench} on 5s short video generation.}
\label{tab:short-video-gen}
\setlength{\tabcolsep}{5pt}
\resizebox{\linewidth}{!}{%
\begin{tabular}{c cc ccc}
\toprule
Direction & Shared $G^\theta$ & Shared $\mu_\text{fake}^\phi$ & Quality $\uparrow$ & Semantics $\uparrow$ & Total $\uparrow$ \\
\midrule
\mr{3}{$\rightarrow$}
  & \xmark & \xmark & 84.47 & 79.23 & 83.42 \\
  & \cmark & \xmark & 84.64 & 79.18 & 83.54 \\
  & \cmark & \cmark & \textbf{84.89} & \textbf{79.51} & \textbf{83.81} \\
\midrule
\mr{3}{$\leftarrow$}
  & \xmark & \xmark & \textbf{85.17} & 80.61 & 84.26 \\
  & \cmark & \xmark & 85.02 & 79.74 & 83.96 \\
  & \cmark & \cmark & 85.12 & \textbf{80.69} & \textbf{84.23} \\
\bottomrule
\end{tabular}
}%
\end{minipage}
\end{table*}

\noindent\textbf{Single-direction short video generation.}
\cref{tab:short-video-gen} reports VBench results on 5s video generation.
The configuration with separate $G^\theta$ and separate $\mu_\text{fake}^{\phi}$ (first row) corresponds to the Self-Forcing~\cite{selfforcing} baseline.
In comparison, our unified method achieves competitive performance in both directions compared to the Self-Forcing baseline. We also observe slightly higher performance in backward generation. This can be attributed to learning asymmetry rooted in forward bias in the latent space.

\begin{table}[t!]
\centering
\caption{\textbf{Long video generation evaluated with VBench~\cite{vbench}}. We compare forward-only baselines with \method in both directions under various sink latent configurations.
  Past/future sink latents provide temporal anchoring to control content variation over long durations.}
\label{tab:long-video-gen}
\setlength{\tabcolsep}{3.5pt}
\resizebox{\textwidth}{!}{%
\begin{tabular}{cl cc l ccccccc}
\toprule
Direction & Method & Past Sink & Future Sink & Dur. & Subj. $\uparrow$ & BG $\uparrow$ & Flicker $\uparrow$ & Smooth $\uparrow$ & Dynamic $\uparrow$ & Aes. $\uparrow$ & Img. $\uparrow$ \\
\midrule
\mr{18}{$\rightarrow$}
  & \mr{3}{LongLive~\cite{longlive2025}}
    & \mr{3}{\cmark} & \mr{3}{\xmark}
      & 10s  & 96.84 & 96.26 & \textbf{97.77} & \textbf{98.84} & 36.72 & 62.53 & 70.65 \\
  & & & & 30s  & 96.45 & 95.45 & \textbf{97.65} & \textbf{98.75} & 38.28 & 61.12 & 69.20 \\
  & & & & 100s & 96.48 & 95.52 & \textbf{97.71} & \textbf{98.76} & 36.72 & 61.81 & 69.22 \\
  \cmidrule(l){2-12}
  & \mr{3}{Rolling-Forcing~\cite{rollingforcing2025}}
    & \mr{3}{\cmark} & \mr{3}{\xmark}
      & 10s  & 96.78 & 95.48 & 97.65 & 98.70 & 28.12 & 62.03 & 71.16 \\
  & & & & 30s  & 96.60 & 95.37 & \textbf{97.65} & 98.72 & 31.25 & 60.98 & \textbf{70.96} \\
  & & & & 100s & 96.06 & 95.19 & 97.45 & 98.57 & 29.69 & 58.96 & 70.99 \\
  \cmidrule(l){2-12}
  & \mr{3}{Infinity-Rope~\cite{infinityrope2025}}
    & \mr{3}{\cmark} & \mr{3}{\xmark}
      & 10s  & 95.48 & 94.74 & 96.29 & 97.93 & 64.06 & 61.17 & 70.18 \\
  & & & & 30s  & 95.11 & 94.63 & 96.13 & 97.82 & 59.38 & 58.78 & 69.38 \\
  & & & & 100s & 94.93 & 94.53 & 96.05 & 97.74 & 55.47 & 56.26 & 69.39 \\
  \cmidrule(l){2-12}
  & \mr{3}{Self-Forcing~\cite{selfforcing}}
    & \mr{3}{\cmark} & \mr{3}{\xmark}
      & 10s  & 96.47 & 95.36 & 96.36 & 98.31 & 64.06 & 62.74 & \textbf{71.23} \\
  & & & & 30s  & 95.75 & 95.01 & 95.88 & 98.08 & 64.06 & 59.84 & 70.33 \\
  & & & & 100s & 96.14 & 94.84 & 96.01 & 98.12 & 62.50 & 56.85 & 70.75 \\
  \cmidrule(l){2-12}
  & \mr{3}{\method}
    & \mr{3}{\cmark} & \mr{3}{\xmark}
      & 10s  & 95.40 & 94.37 & 95.11 & 97.69 & \textbf{81.25} & 61.64 & 70.22 \\
  & & & & 30s  & 94.36 & 93.54 & 94.53 & 97.43 & \textbf{92.97} & 57.70 & 69.10 \\
  & & & & 100s & 93.95 & 93.23 & 94.43 & 97.37 & \textbf{87.50} & 56.55 & 69.53 \\
  \cmidrule(l){2-12}
  & \mr{3}{\method}
    & \mr{3}{\cmark} & \mr{3}{\cmark}
      & 10s  & \textbf{97.60} & \textbf{96.30} & 97.50 & 98.50 & 46.80 & 62.70 & 70.70 \\
  & & & & 30s  & \textbf{97.40} & \textbf{96.00} & 97.00 & 98.10 & 44.50 & \textbf{62.90} & 70.80  \\
  & & & & 100s & \textbf{97.30} & \textbf{95.80} & 96.90 & 98.00 & 44.50 & 62.30 & \textbf{71.30} \\
\midrule
\mr{6}{$\leftarrow$}
  & \mr{3}{\method}
    & \mr{3}{\xmark} & \mr{3}{\cmark}
      & 10s  & 94.91 & 94.57 & 95.94 & 98.10 & 71.09 & \textbf{63.35} & 70.38 \\
  & & & & 30s  & 93.89 & 93.67 & 94.93 & 97.73 & 81.25 & 60.31 & 70.71 \\
  & & & & 100s & 93.94 & 93.50 & 94.67 & 97.64 & 85.16 & 59.02 & 70.51 \\
  \cmidrule(l){2-12}
  & \mr{3}{\method}
    & \mr{3}{\cmark} & \mr{3}{\cmark}
      & 10s  & 96.87 & 96.03 & 97.13 & 98.34 & 55.47 & 62.95 & 70.39 \\
  & & & & 30s  & 96.80 & 95.60 & 97.04 & 98.28 & 60.16 & 62.69 & 70.49 \\
  & & & & 100s & 96.75 & 96.00 & 97.04 & 98.28 & 56.25 & \textbf{62.83} & 70.31 \\
\bottomrule
\end{tabular}%
}\vspace{-1em}
\end{table}

\begin{figure}[t!]
\centering
\includegraphics[width=\linewidth]{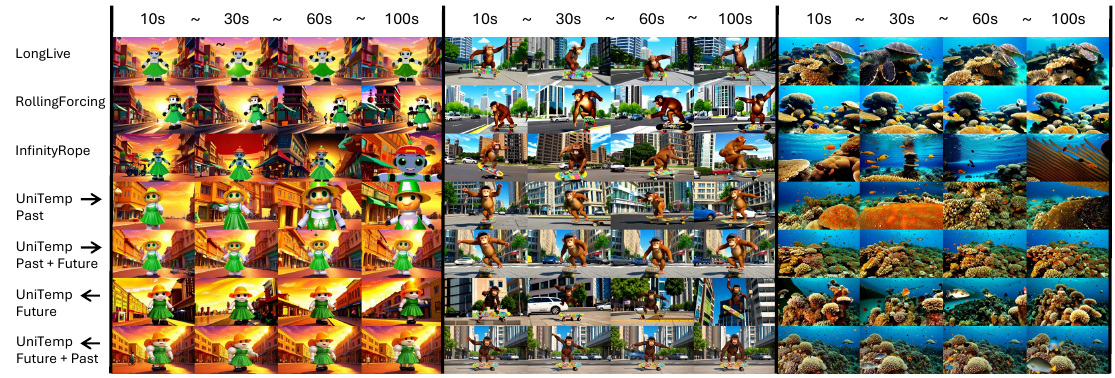}
\caption{\textbf{Long video generation results}. Past + future sink latents provide strong conditioning to reduce content variation over long durations.}
\label{fig:long-video}
\end{figure}

\FloatBarrier

\medskip
\noindent\textbf{Single-direction long video generation.}
\cref{tab:long-video-gen} compares long video generation at 10s, 30s, and 100s.
Existing training-based long video generation methods (LongLive~\cite{longlive2025}, Rolling-Forcing~\cite{rollingforcing2025}) achieve high temporal consistency but produce extremely low-dynamic content ($36.72/28.12$).
Note that this is a tradeoff in VBench metrics: more dynamic content usually leads to lower temporal consistency, because there are larger differences between neighboring frames.
In comparison, training-free methods with past sink latents~\cite{infiniteforcing2025} can already achieve strong consistency with slightly degraded dynamics and aesthetic quality.
E.g., from 10s to 100s, Infinity-Rope's dynamic degree / aesthetic quality drops from $64.06/61.17$ to $55.47/56.26$, while Self-Forcing's dynamic degree / aesthetic quality decreases from $64.06/62.74$ to $62.74/56.85$.
Our \method inherits such stable long video generation capability but with increased dynamic degree, which leads to lower frame consistency.
However, we can control such dynamics-consistency tradeoff by switching between single-end sink latents and both-ends sink latents.
In forward generation, by enabling both-end-guided sink latents, the generated content faithfully follows the sink frames,
  effectively reducing drifting from 10s to 100s with (1) stable and high subject consistency ($97.60\rightarrow 97.30$) (2) lower but stable dynamic degree ($46.80\rightarrow 44.50$) and (3) stable aesthetic quality ($62.70\rightarrow 62.30$).
Backward generation shows similar patterns.
Furthermore, we can see the visual results in \cref{fig:long-video}, where \method can generate stable long videos with content variations controlled by single / both sink latents.

\medskip
\noindent\textbf{Inbetween video generation}
\cref{tab:inbetween-video-gen} evaluates inbetween generation given the first and last frames.
Wan FLF2V~\cite{wan2024} is a large 14B-parameter model with high computational cost that has been trained for inbetween generation tasks on a large amount of video data.
GI is built on Stable Video Diffusion~\cite{stablevideodiffusion2024} with a fine-tuned backward image-to-video diffusion model. At inference, the denoising steps need to be run for both forward and backward image-to-video generation, leading to inefficiency in both memory and speed.
Both Wan FLF2V and GI support only a single frame on each side for conditioning.

In comparison, \method is not specifically trained for inbetween generation tasks and has only one unified model. The shared architecture allows simultaneously conditioning on both head and tail frames, while its autoregressive nature allows conditioning on flexible numbers of head and tail frames.
\method shows strong performance by outperforming GI and Wan FLF2V in background consistency, image quality, and video distribution.
Different from GI and Wan FLF2V, \method can reuse latents of conditioning frames and thus does not need to reproduce the conditioning frames, leading to much higher boundary consistency.
Qualitatively, as shown in \cref{fig:inbetween}, GI often assumes linear motion between head and tail frames, resulting in sometimes unnatural content and sudden blending effects,
especially when the head and tail frames are far apart (the second case).
In contrast, Wan FLF2V and \method can generate intermediate content with natural variations, while \method is much more efficient in both training and inference.

\begin{table}[t]
\centering
\caption{\textbf{Inbetween video generation evaluation}. We compare \method against Wan FLF2V~\cite{wan2024} and GI~\cite{gi2024} on bounded generation given head and tail frames. Head/tail flickering measures boundary consistency with the ground-truth frames.}
\label{tab:inbetween-video-gen}
\setlength{\tabcolsep}{3.5pt}
\resizebox{\textwidth}{!}{%
\begin{tabular}{l ccccccc cc cc}
\toprule
& \multicolumn{7}{c}{VBench Dimensions} & \multicolumn{2}{c}{Boundary} & \multicolumn{2}{c}{Distribution} \\
\cmidrule(lr){2-8} \cmidrule(lr){9-10} \cmidrule(lr){11-12}
Method & Subj. $\uparrow$ & BG $\uparrow$ & Flicker $\uparrow$ & Smooth $\uparrow$ & Dynamic $\uparrow$ & Aes. $\uparrow$ & Img. $\uparrow$ & Head $\uparrow$ & Tail $\uparrow$ & FID $\downarrow$ & FVD $\downarrow$ \\
\midrule
GI~\cite{gi2024}    & \textbf{94.48} & 92.90 & 97.27 & \textbf{98.75} & 62.50 & 60.19 & 62.88 & 96.29 & 97.47 & 34.21 & 34.66 \\
Wan FLF2V~\cite{wan2024}           & 93.44 & 93.63 & \textbf{97.38} & 98.54 & \textbf{64.84} & 64.38 & 64.59 & 98.88 & \textbf{98.52} & \textbf{23.31} & 27.77 \\
\method             & 93.81 & \textbf{93.71} & 97.18 & 98.46 & 63.28 & \textbf{65.75} & \textbf{66.68} & \textbf{99.52} & 98.16 & 24.94 & \textbf{25.00} \\
\midrule
Ground-truth                 & 94.11 & 94.78 & 97.64 & 98.66 & 57.81 & 61.17 & 64.25 & 100.00 & 100.00 & -- & -- \\
\bottomrule
\end{tabular}%
}
\end{table}

\begin{figure}[t]
\centering
\includegraphics[width=\linewidth]{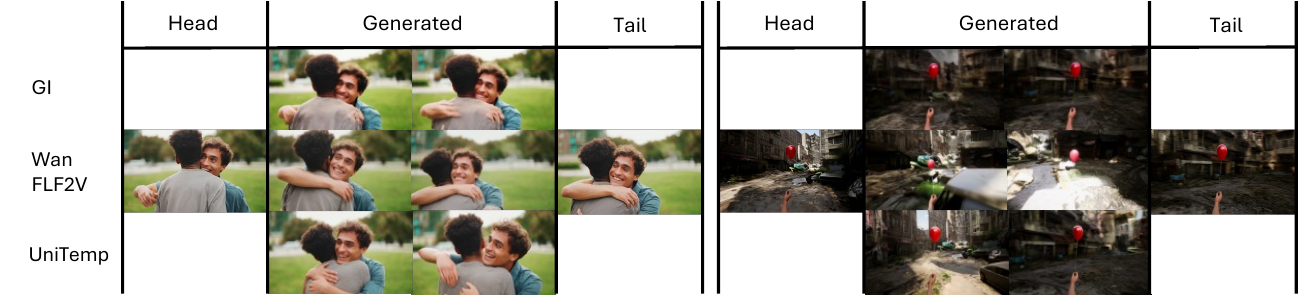}
\caption{Visualization of inbetween video generation. Given the head (leftmost) and tail (rightmost) frames, \method infills temporally coherent content.}\vspace{-1em}
\label{fig:inbetween}
\end{figure}

\subsection{Ablation Studies}\label{sec:experiments:ablation}

\noindent\textbf{Effect of anchor latents}
\cref{tab:inter-block-flickering} reports inter-block and intra-block flickering ratio for varying numbers of anchor latents $P$.
The baseline without anchor latents exhibits a backward inter-block flickering ratio of $1.42$, confirming substantial boundary artifacts.
Increasing $P$ progressively reduces this: $P{=}1$ ($1.26$), $P{=}2$ ($1.23$), $P{=}3$ ($\mathbf{1.07}$).
At $P{=}3{=}B$, the ratio approaches the ideal value of $1.0$.
In addition, we investigate whether the anchor latents should be included as outputs in training and inference.
When included as outputs in training, the loss is applied to the anchor latents. As discussed in \cref{sec:method:anchor}, anchor latents themselves are generated without past context.
Therefore, once included in the outputs at test time, they are positioned near block boundaries and still show inter-block flickering, as validated in \cref{tab:inter-block-flickering} ($1.07 \rightarrow 1.46$, $1.08 \rightarrow 1.48$). We show video visualizations of this flickering in the appendix.

\medskip
\noindent\textbf{Unified vs.\ separate models.}
\cref{tab:short-video-gen} compares three training configurations.
Overall, training direction-specific separate models for either the student generator $G^\theta$ or fake diffusion model $\mu_\text{fake}^{\phi}$ does not bring significant performance gains but introduces higher training cost.
Our unified training framework can also reduce the performance difference between forward ($84.89/79.51/83.81$) and backward ($85.12/80.69/84.23$) generation.
This demonstrates the effectiveness and efficiency of our unified training framework.

\subsection{Applications}\label{sec:experiments:application}
Since \method enables flexible conditioning at inference time, we demonstrate its effectiveness across several applications. Implementation details for these applications are provided in the supplementary material.

\begin{figure}[t]
\centering
\begin{minipage}{0.48\linewidth}
    \centering
    \includegraphics[width=\linewidth]{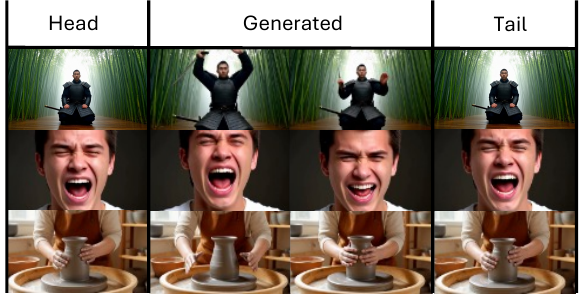}
    \caption{Looping video generation given the same head and tail frames.}
    \label{fig:looping}
\end{minipage}
\hfill
\begin{minipage}{0.48\linewidth}
    \centering
    \includegraphics[width=\linewidth]{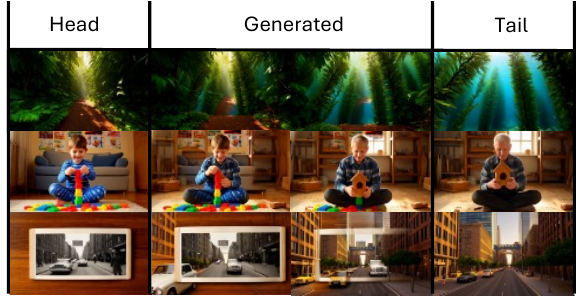}
    \caption{Scene transition given distinct head and tail frames.}
    \label{fig:scene-transition}
\end{minipage}\vspace{-1em}
\end{figure}

\medskip
\noindent\textbf{Looping video generation.}
By shifting the head conditioning frames to the future and then generating the intermediate content between them, we can easily achieve looping video generation,
as shown in \cref{fig:looping}.

\medskip
\noindent\textbf{Scene transition.}
When the head and tail conditioning frames are very different, we face the challenging task of scene transition.
Surprisingly, we find that \method can perform reasonably well in this task between contrasting scenes. As shown in \cref{fig:scene-transition},
  \method is able to associate visually similar content and apply a smooth transition.

\medskip
\noindent\textbf{Long-shot visual story generation.}
With versatile generation orders (forward, backward, inbetween) and conditioning frames, we are now able to build visual stories in any order we want, instead of only extending forward.
For example, in \cref{fig:teaser}, we can generate multiple key scenes first, then flexibly extend forward / backward / infill between any two existing scenes, allowing looping long-shot visual story generation across different numbers of scenes and durations.

\section{Conclusion and Discussion}\label{sec:conclusion}

We presented \method, a unified autoregressive video generation framework that supports generation in any temporal order.
We identified the cross-block flickering issue in backward-generated videos, which is associated with the causal temporal bias of Causal 3D VAE latents. We proposed blockwise anchor latents to help restore missing past context at block boundaries.
Built on a unified distillation framework, \method trains a single student model for both forward and backward generation. Computed features can then be shared across any-direction generation workflows, enabling efficient inference.
The resulting model achieves competitive performance on short and long video generation as well as inbetween generation, while unlocking applications such as looping video, scene transitions, and visual story generation, without task-specific fine-tuning.

\smallskip
\noindent\textbf{Limitations.}
Backward generation incurs additional per-block computation due to the jointly denoised anchor latents; forward generation is unaffected.
The causal VAE and teacher model remain inherently forward-biased: anchor latents mitigate but do not fully eliminate this asymmetry.
Exploring bidirectional VAE architectures that maintain compatibility with existing pre-trained teachers is a promising direction for future work.

\bibliographystyle{style-eccv/splncs04}
\bibliography{refs}
\endgroup

\clearpage

\setcounter{section}{0}
\renewcommand{\thesection}{\Alph{section}}

\begin{center}
{\Large\bfseries UniTemp: Unlocking Video Generation in Any Temporal Order via Bidirectional Distillation\\[0.3em]
Supplementary Material\par}
\end{center}
\medskip

We use numbers (\eg, Sec.\ 1) to refer to the main paper and capital letters (\eg, Sec.\ A) to refer to this supplement.

\medskip
\setcounter{tocdepth}{2}
\begingroup
\renewcommand{\contentsname}{}
\let\clearpage\relax
\tableofcontents
\endgroup

\section{Implementation Details}\label{sec:appendix-impl}

\subsection{Training Details}

\noindent\textbf{Stage-1 training.}
In the first stage, we initialize the student model by training it to approximate the teacher's ODE trajectories.
We use the Wan2.1 T2V 14B~\cite{wan2024} teacher model to generate 16K ODE trajectory pairs with a timestep shift of $5.0$ and classifier-free guidance scale of $3.0$ from the VidProm~\cite{vidprom} dataset.
The student model is initialized from the pretrained Wan2.1 T2V 1.3B and trained with the AdamW optimizer with
a learning rate of $2e-6$, $\beta_1 = 0.9$, $\beta_2 = 0.999$, weight decay $= 0.01$, a total batch size of 128 across 8x8 H100/A100 GPUs, for a total of 6K iterations.
Denoising steps: $1000/937.50/833.33/625.00$.
We use mixed precision (bfloat16) training with gradient checkpointing and FSDP.
Both forward and backward attention masks are applied in each training iteration to enable bidirectional training.
Block size is $B = 3$ with anchor latent size $P = 3$.
We train with a diffusion forcing schedule, i.e., independently sampling denoising steps $t_i$ for each block in one sample, following the implementation in Self-Forcing~\cite{selfforcing}.

\medskip
\noindent\textbf{Stage-2 training.}
In the second stage, we perform distribution matching distillation (DMD)~\cite{dmd,dmdv2} to further train the student model for autoregressive generation.
The student model is initialized from the Stage-1 checkpoint.
We still use the pretrained Wan2.1 T2V 14B as the real score network, which is kept frozen during training.
For the generator (student), we use the AdamW optimizer with a learning rate of $2e-6$, $\beta_1 = 0.0$, $\beta_2 = 0.999$, weight decay $= 0.01$.
For the fake critic model, we use the AdamW optimizer with a learning rate of $4e-7$, $\beta_1 = 0.0$, $\beta_2 = 0.999$, weight decay $= 0.01$.
The total batch size is 64 across 8x8 H100/A100 GPUs.
We train for 600 iterations and apply exponential moving average (EMA) with a decay rate of $0.99$ starting from iteration 200 on the generator.
Training is \textbf{video data free}: we only use text prompts from VidProm~\cite{vidprom} (rewritten by Qwen2.5-7B-Instruct~\cite{qwen25instruct}) without any video data.
The denoising schedule is kept the same as in stage-1, i.e., $[1000.00,  937.50,  833.33,  625.00]$ (4 steps) with a timestep shift of $5.0$.
When training the fake critic model, we freeze the student model and ask the fake critic model to approximate videos generated by the frozen student model in both forward and backward directions.
When training the student model (generator), we freeze the fake critic model, and send the latents generated by the student to both the fake critic model and the real score teacher network to compute the DMD gradients.
We train the student model every 5 steps and the fake critic model every step.

\subsection{Evaluation Details}

\noindent\textbf{Flickering ratio.}
To evaluate inter-block flickering artifacts that are more pronounced in high-dynamic scenes, we use an LLM to generate 128 high-dynamic video prompts describing scenes with significant motion (\eg, fast-moving objects, dynamic camera movements, action sequences).
For each generated video, we compute the flickering ratio (FR) as defined in Eq.~(1) of the main paper.
We split the FR measurements into two groups:
\begin{itemize}[nosep]
  \item \textbf{Inter-block FR}: computed between adjacent latents $z_i$ and $z_{i+1}$ that belong to \emph{different} generated blocks, \ie at block boundaries.
  \item \textbf{Intra-block FR}: computed between adjacent latents $z_i$ and $z_{i+1}$ that belong to the \emph{same} generated block.
\end{itemize}
We average the FR values within each group across all prompts. An ideal model should have both inter-block and intra-block FR close to $1.0$, indicating uniform temporal smoothness regardless of block boundaries.

\medskip
\noindent\textbf{Single-direction short video generation.}
We follow the standard VBench~\cite{vbench} evaluation protocol.
We generate 5-second videos (81 pixel frames, 21 latent frames) using 946 official VBench prompts rewritten by Qwen2.5-7B-Instruct~\cite{qwen25instruct}, following Self-Forcing~\cite{selfforcing}.
All metrics are computed using the official VBench codebase with default settings. We generate 5 samples per prompt and report averaged scores across 16 VBench quality and semantic dimensions.
To ensure fair comparison, we reimplement the Self-Forcing~\cite{selfforcing} baseline using an identical stage-1 and stage-2 training protocol but with forward-only generation. Note that our training protocol is identical to the original Self-Forcing except
for the stage-1 sampled ODE pairs, which are not disclosed in Self-Forcing.

\medskip
\noindent\textbf{Long video generation.}\label{sec:imp-details:long-video}
Following previous practice~\cite{rollingforcing2025,infinityrope2025,selfforcing++} on long video generation evaluation, we randomly sample 128 prompts from MovieGenBench~\cite{moviegenbench} and generate long videos at durations of 10s, 30s, and 100s for each prompt.
When evaluating both the baseline Self-Forcing and our \method, we follow the training-free techniques introduced in~\cite{infiniteforcing2025}, i.e., treating the initially generated block as sink latents and applying RoPE~\cite{rope} re-indexing after retrieving KV cache.
Note that InfinityRope~\cite{infinityrope2025} also follows these techniques.
In practice, besides always attending to the sink latents, each block also attends to its neighboring generated latents with a local attention window of 6 latents.
When applying future sink latents, we shift the second generated block to future positions with a distance of 3 latent positions via RoPE re-indexing, simulating future conditioning.
Note that the concept of past and future sink latents becomes diffenrence for backward generation,
where conventional sink latents now becomes the future latents, while the past sink latents are placed with a distance of 3 latents before the current block.
We compare to the following baselines:
\begin{itemize}[nosep]
  \item \textbf{LongLive}~\cite{longlive2025}: A training-based method that extends Self-Forcing for real-time interactive long video generation. It fine-tunes a short-clip autoregressive model with additional training on long video sequences, achieving high temporal consistency at the cost of reduced content dynamics.
  \item \textbf{Rolling-Forcing}~\cite{rollingforcing2025}: A training-based method that stabilizes long video generation by denoising frames together in a rolling window for mutual refinement, breaking the chain of error accumulation in standard autoregressive generation.
   It also adopts attention sink mechanism that anchors the video to preserve global context provided by the initial sink latents.
  \item \textbf{Infinity-Rope}~\cite{infinityrope2025}: A training-free method that enables infinite-length video generation by extending RoPE~\cite{rope} temporal indices beyond the training horizon.
  It uses past sink latents with re-indexed temporal positions for long-term stability.
\end{itemize}

\medskip
\noindent\textbf{Inbetween video generation.}
Our inbetween generation evaluation follows this protocol:
\begin{enumerate}[nosep]
  \item \textbf{Ground truth generation}: We use the Wan2.1 T2V 14B teacher model to generate 5-second videos (81 pixel frames $x_{0\rightarrow 80}$ and 21 latent frames $z_{0\rightarrow 20}$) as ground truth, using the same 128 prompts in \cref{sec:imp-details:long-video} from MovieGenBench~\cite{moviegenbench}.
  \item \textbf{Condition extraction}: We keep the first 9 pixel frames and their corresponding 3 latent frames ($z_0, z_1, z_2$) as the head condition, and the last 12 pixel frames and their corresponding 3 latent frames ($z_{18}, z_{19}, z_{20}$) as the tail condition.
  The model is tasked with generating intermediate frames. The number of generated frames depends on the method.
  \item \textbf{\method inference}:
  The baseline methods only support single-frame conditioning at each end, so we feed in $x_8$ and $x_{69}$ as the head and tail condition frames.
  For \method, we condition on 3 head ($z_0, z_1, z_2$) and 3 tail ($z_{18}, z_{19}, z_{20}$) latent frames. We first compute KV cache for the input head latents. For the tail latents, their KV caches are computed by additionally attending to the head latents' KV cache.
  The 15 intermediate latents are then jointly denoised as a single large block with full-sequence self-attention, while attending to the KV cache of both head and tail blocks.
  The final output video is composed by concatenating the latent sequence of (head latents, generated intermediate latents, tail latents) and decoding it into a video via the Causal VAE decoder. We extract frames $8 \rightarrow 69$ (62 frames) for evaluation and comparison with baseline methods.
\end{enumerate}

We evaluate the complete video sequence (head frames, generated intermediate frames, tail frames) using VBench~\cite{vbench} metrics.
Additionally, we compare FID and FVD between the generated video and the ground truth video to evaluate distribution similarity.
Notice that Wan FLF2V and GI both regenerate the conditioning frames, while \method only decode conditioning latents into frames.
To evaluate consistency at boundaries, we also compute the temporal flickering (\ie, $1-$ average pixel L1 distance) between the groundtruth head/tail frames and the generated head/tail frames.
We compare to the following baselines:
\begin{itemize}[nosep]
  \item \textbf{Wan FLF2V}~\cite{wan2024}: A large 14B-parameter model that has been trained for inbetween generation tasks on a large amount of video data.
  \item \textbf{Generative Inbetweening (GI)}~\cite{gi2024}: Built on Stable Video Diffusion~\cite{stablevideodiffusion2024}, GI adapts an image-to-video diffusion model for keyframe interpolation by training a separate backward generation branch.
  At inference, it runs both forward and backward image-to-video generation from the two conditioning frames and blends the results via weighted averaging.
  This requires running the full denoising process twice (once per direction), leading to higher computational cost.
  GI only supports single-frame conditioning at each end and generates 25 intermediate frames at $1024 \times 576$ resolution, which are resized to the target resolution for evaluation.
\end{itemize}

\section{Anchor Latents Details}\label{sec:appendix-anchor}

We describe how backward generation with anchor latents is implemented during Stage-1 training, Stage-2 training, and inference, and discuss several alternative design choices.
We also include the detailed attention masks and generation orders in \cref{fig:vis-attn-mask} and \cref{fig:vis-gen-order}, respectively.
Finally, in \cref{sec:anchor-details:ablation}, we present ablation results analyzing the effects of these design choices across the three stages.

\subsection{Stage-1 Training}\label{sec:anchor-details:stage-1-training}

\begin{figure}[ht!]
\centering
\begin{tabular}{@{}c@{\hspace{4pt}}c@{}}
\includegraphics[width=0.48\linewidth]{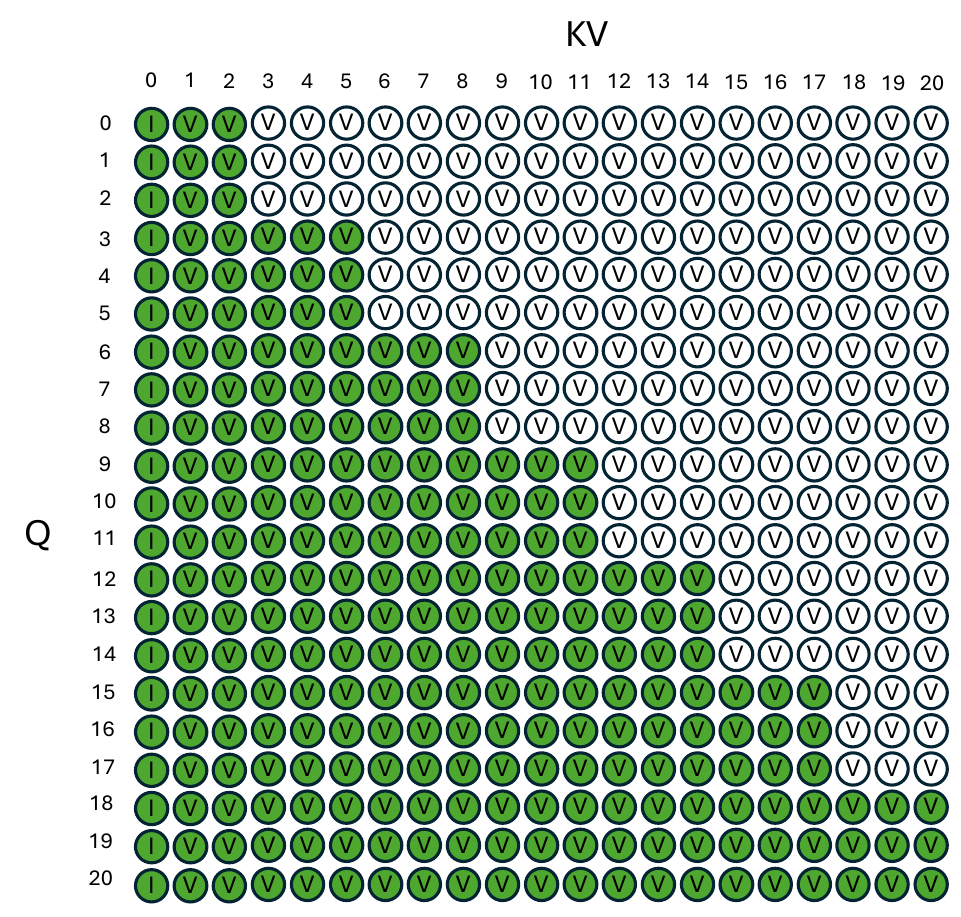} &
\includegraphics[width=0.48\linewidth]{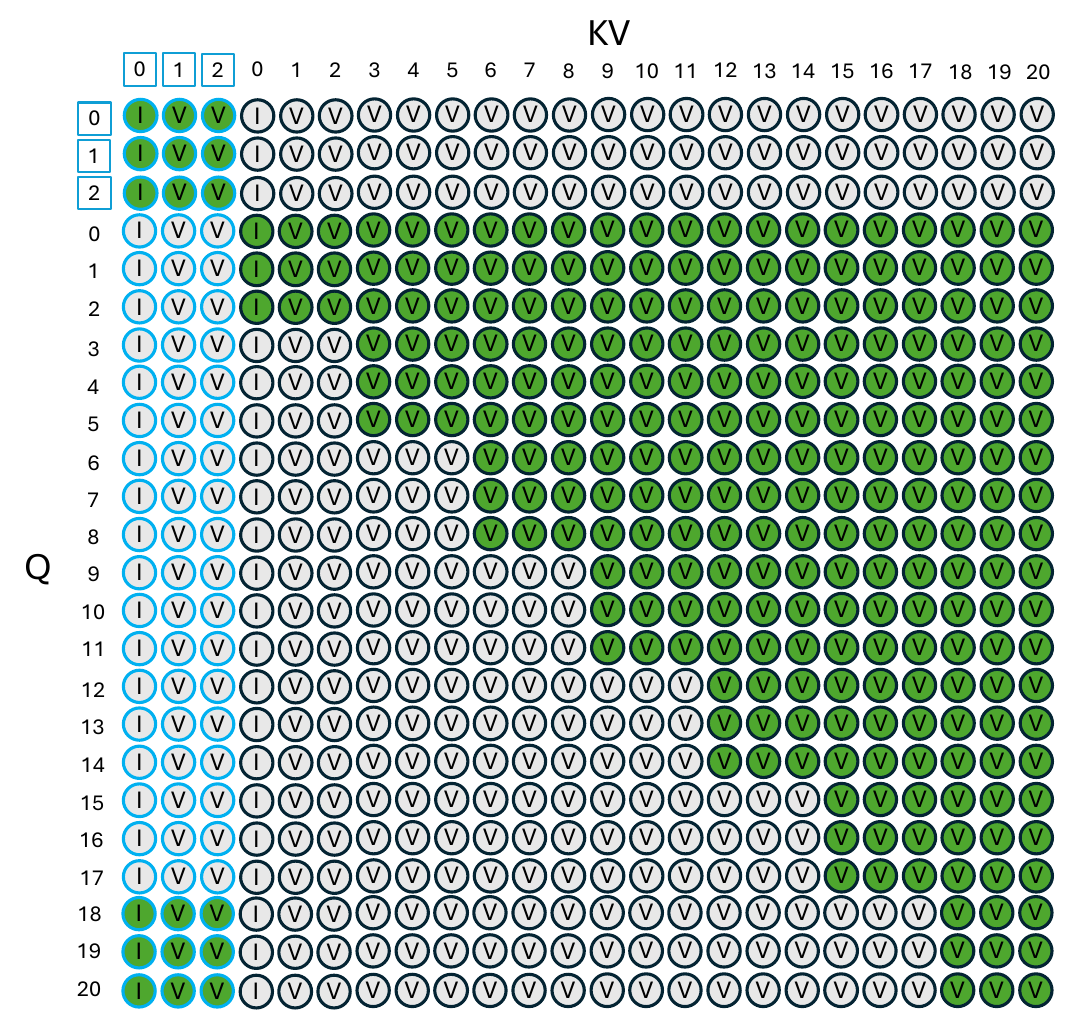} \\
\parbox{0.48\linewidth}{\centering (a) Forward} &
\parbox{0.48\linewidth}{\centering (b) Baseline Backward ($B{=}3$, no anchor)} \\[6pt]
\includegraphics[width=0.48\linewidth]{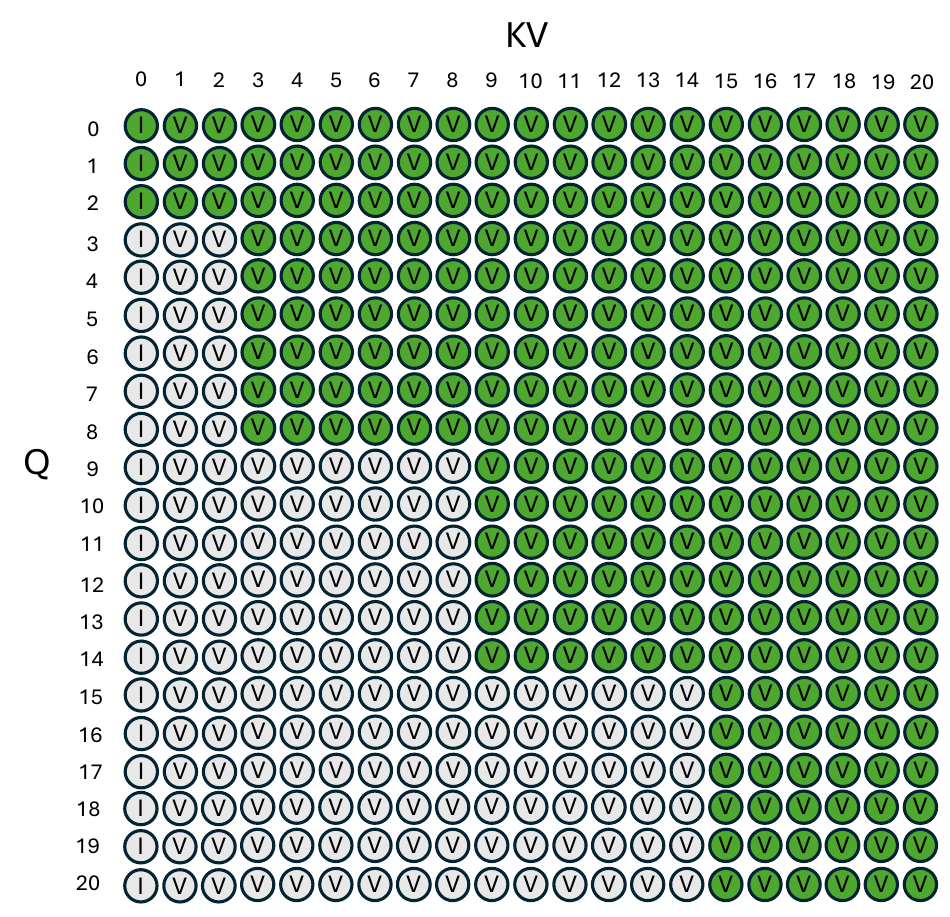} &
\includegraphics[width=0.48\linewidth]{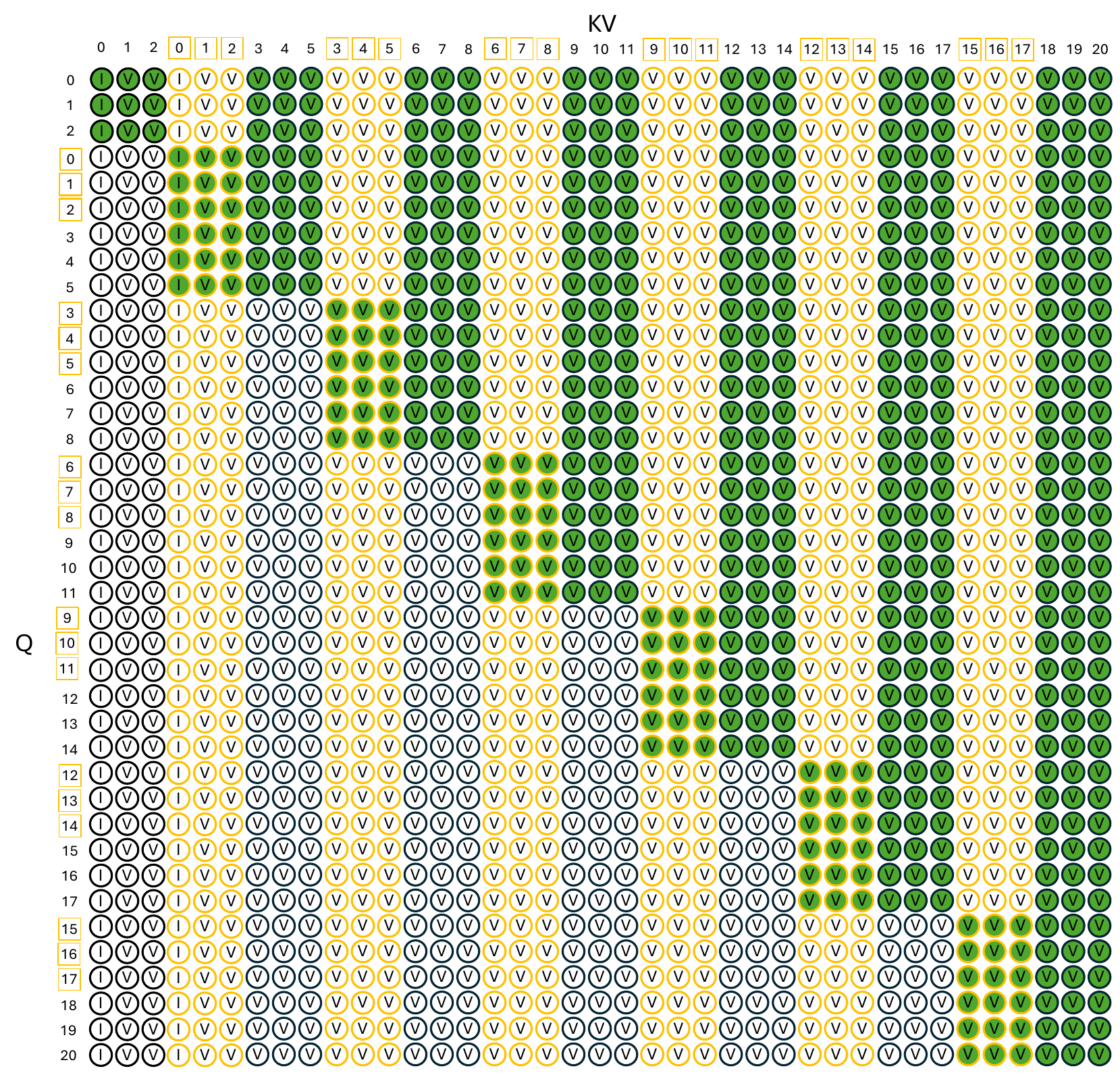} \\
\parbox{0.48\linewidth}{\centering (c) Baseline Backward ($B{=}6$, no anchor)} &
\parbox{0.48\linewidth}{\centering (d) \method Backward ($B{=}P{=}3$, with anchor)} \\
\end{tabular}
\caption{\textbf{Attention masks for Stage-1 training.} Attended latents are filled with \textcolor{ForestGreen}{green} color.
(a) Forward causal mask: each block attends to all previously generated blocks.
(b) Baseline backward attention mask with block size $B{=}3$ and without anchor latents: each block attends to future blocks, while we introduce a dummy initial block (shown in \textcolor{cyan}{blue}) to resolve the image/video latent ambiguity.
(c) Baseline backward attention mask with block size $B{=}6$ and without anchor latents:
we generate 6 latents in each block in backward order. A dummy initial block is not necessary since the model can distinguish image/video latents by the number of latents in the first generated block in forward ($B{=}3$) and backward ($B{=}6$) generation.
(d) Our \method backward attention mask with block size $B{=}3$ and anchor size $P{=}3$: we introduce anchor latents (shown in \textcolor{yellow}{yellow}) to be jointly denoised with each block, serving as past context to stabilize backward generation and resolve the inter-block flickering issue.
Anchor latents share the same noise level as the corresponding generated latents, while different blocks have different noise levels to learn diffusion forcing. Zoom in to check the frame indices of the anchor latents.
}
\label{fig:vis-attn-mask}\vspace{-2em}
\end{figure}

In Stage-1 (ODE initialization), we apply both forward and backward attention masks for bidirectional autoregressive training.
Within each block, all input latents share the same noise level, and can mutually attend to each other in all attention layers.
The noise levels for different blocks are sampled independently.
Throughout all our experiments, we always use a block size of 3 for forward generation. Backward block size is set to $B$ in baseline methods, or $P+B$ when anchor latents are used.
For ablation, we also tried simply increasing the block size in backward generation to $P+B$ without using anchor latents, as shown in \cref{fig:vis-attn-mask}~(c).
By default, we set both $B$ and $P$ to 3.
We show the attention masks in \cref{fig:vis-attn-mask}, where (a) and (d) are our final unified attention masks used in training for forward and backward generation, respectively.

Note that in the baseline method~(b), when we use block size $B=3$ in both forward and backward generation, a conflict arises on the first block's generation:
the forward pass generates its first block ($z_0, z_1, z_2$) where $z_0$ is a special image latent, while the backward pass generates its first block ($z_{18}, z_{19}, z_{20}$) where $z_{18}$ is a video latent.
The RoPE mechanism treats these two cases identically. Therefore, the model can get confused about whether the first block should start with an image or a video latent.
To resolve this, during backward generation, we additionally sample a dummy block ($z_{0}^{\emptyset}, z_{1}^{\emptyset}, z_{2}^{\emptyset}$) to condition the generation of the first block ($z_{18}, z_{19}, z_{20}$).
With 6 latents in attention, RoPE can thus distinguish the two cases and allow the model to generate correctly.
Loss is not applied on the dummy block. The noise level is sampled independently for the dummy block and the real initial block ($z_0, z_1, z_2$).
In stage-2 training, we also prepend a dummy block in the baseline model's backward generation, as introduced later in \cref{sec:anchor-details:stage-2-training}.

When the backward block size differs from the forward block size, such a dummy block is unnecessary, because the first block in backward generation contains a different number of latents than in forward generation, e.g.,
($z_{15}, z_{16}, z_{17}, z_{18}, z_{19}, z_{20}$) in backward versus ($z_0, z_1, z_2$) in forward, which helps the model easily distinguish the two cases.
This includes cases when backward block size is $6$ (\cref{fig:vis-attn-mask} (c)) or when anchor latents are included (\cref{fig:vis-attn-mask} (d)).

When anchor latents are introduced in Stage-1 training, they have corresponding ground-truth denoising targets from the sampled ODE pairs.
To allow fast training, we sample from these anchor latents' positions two times:
One for the anchor latents, which is sampled to have the same noise level as the corresponding generated latents, i.e., the $P+B$ latents within one block has the same noise level;
The other one for the actual generated latents in the previous block. Noise levels in different blocks are sampled independently to learn diffusion forcing.
This replication design ensures denoising of one block does not rely on its previous block, thus faithfully following the backward generation order.

In practice, we can choose to apply loss on these anchor latents or not.
In this way, in stage-1 training, we have different options:
\begin{itemize}[nosep]
  \item Do not use anchor latents yet keep the same total block size, as in \cref{fig:vis-attn-mask} (c).
  \item Use anchor latents as in \cref{fig:vis-attn-mask} (d), but do not apply loss on them.
  \item Use anchor latents as in \cref{fig:vis-attn-mask} (d), and apply loss on them.
\end{itemize}
We discuss these options in \cref{sec:anchor-details:ablation}.

\subsection{Stage-2 Training}\label{sec:anchor-details:stage-2-training}

\begin{figure}[ht!]
\centering
\begin{tabular}{@{}cc@{}}
\includegraphics[width=0.48\linewidth]{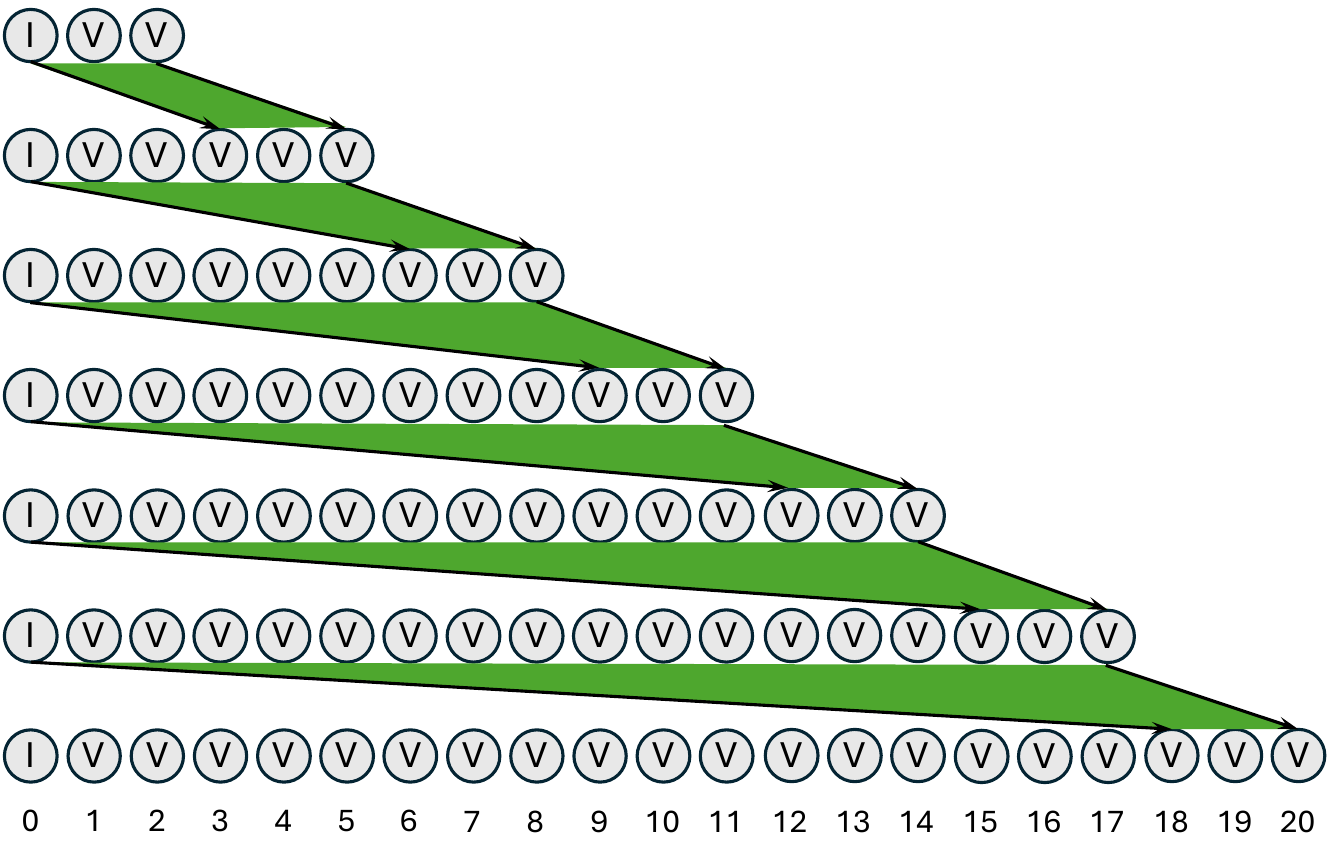} &
\includegraphics[width=0.48\linewidth]{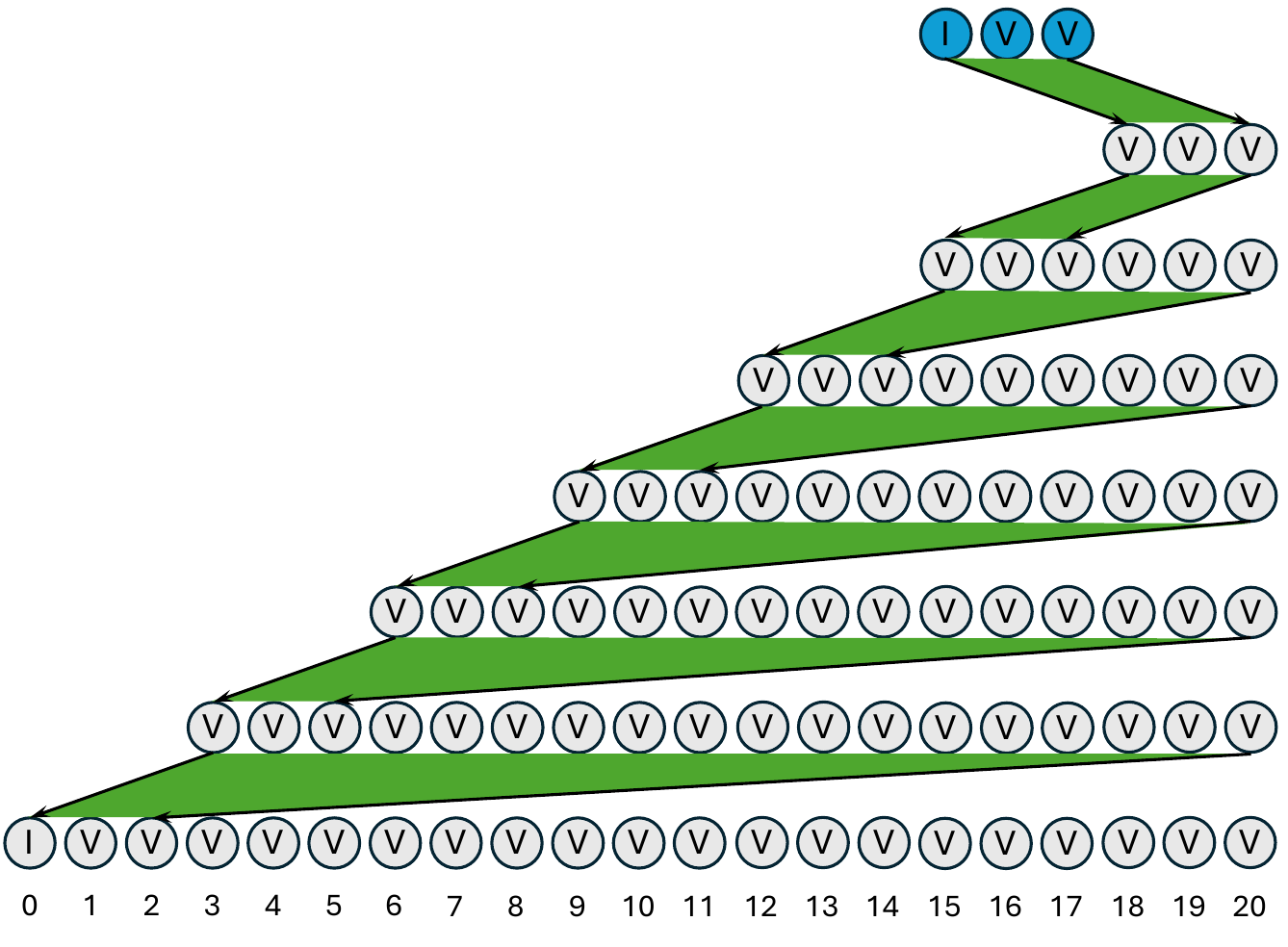} \\
(a) Forward & 
(b) Baseline Backward ($B=3$, no anchor) \\[6pt]
\includegraphics[width=0.48\linewidth]{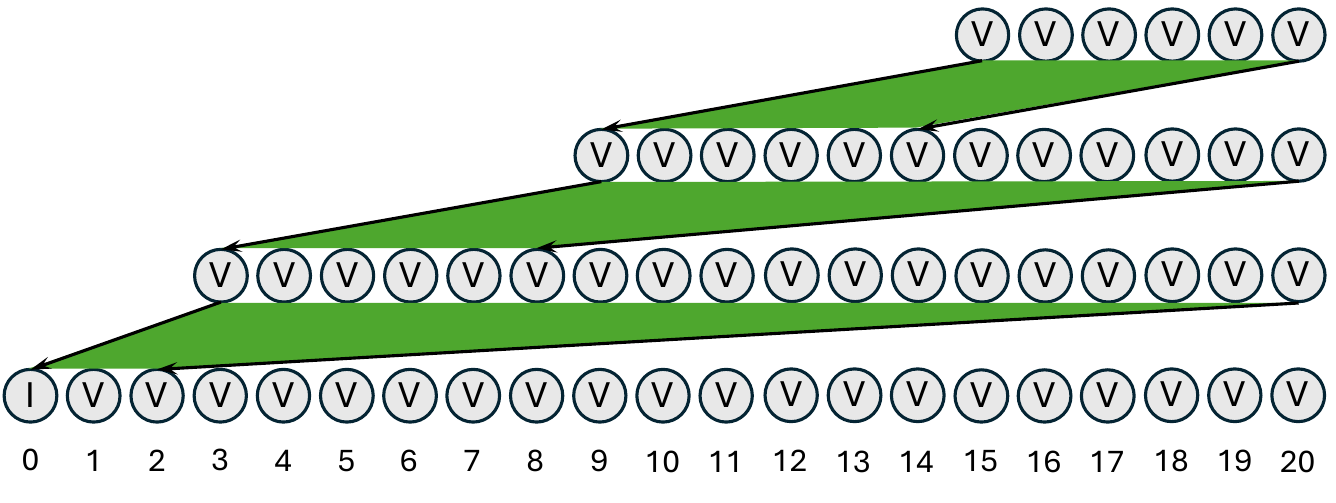} &
\includegraphics[width=0.48\linewidth]{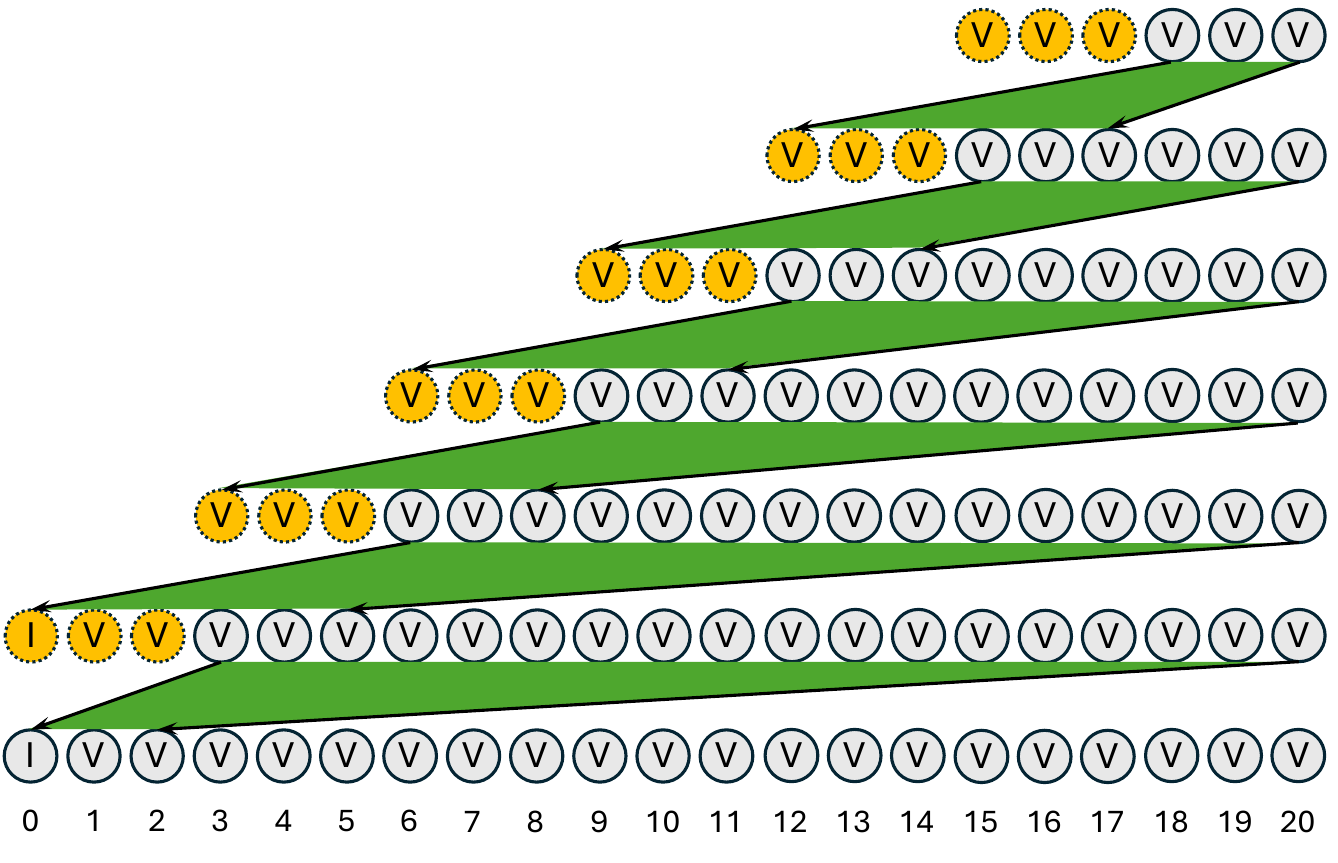} \\
(c) Baseline Backward ($B=6$, no anchor) & 
(d) \method Backward ($B=P=3$) \\
\end{tabular}
\caption{\textbf{Generation order and attended tokens in Stage-2 training.} 
(a) Forward generation: blocks are generated in causal order, each attending to KV cache of all previous blocks. 
(b) Baseline backward generation with block size $B=3$ and without anchor latents: blocks are generated in reverse order, each attending to KV cache of future blocks. A dummy block (shown in \textcolor{cyan}{blue}) is initially generated to serve as a condition to resolve the image/video latent ambiguity.
(c) Baseline backward generation with block size $B=6$ and without anchor latents: blocks are generated in reverse order, each attending to KV cache of future blocks.
(d) Our \method backward generation with block size $B=3$ and anchor size $P=3$: anchor latents (shown in \textcolor{yellow}{yellow}) serve as past context and are jointly denoised within each block. The generated latents can attend to this approximate past context.}
\label{fig:vis-gen-order}
\end{figure}

In Stage-2 training, the student model (generator) generates videos. The entire generated video is supervised by the DMD loss to match the teacher's output distribution.
We can simply reverse the block generation order to simulate backward generation.
We design the generation order and attended positions to match stage-1, as shown in \cref{fig:vis-gen-order}.

As aforementioned, in the baseline method with $B=3$ in both forward and backward generation and without anchor latents, a conflict arises on the first block's generation.
Therefore, in backward generation, before we generate the actual first block, we generate an initial dummy block without gradients. We store the KV cache of this dummy block to condition the generation of the actual first block,
after which the latents and KV cache of the dummy block are discarded, as shown in \cref{fig:vis-gen-order}~(b).

The DMD loss in Stage-2 is applied to the entire video sequence generated by the student, rather than to each latent individually as in Stage-1.
When anchor latents are enabled, the student's generated sequence strictly does not contain these anchor latents.
Therefore, unlike Stage-1 where we can optionally apply loss on the anchor latents, in Stage-2 we cannot apply loss on them.
This gives us two options in stage-2 training:
\begin{itemize}[nosep]
  \item Use anchor latents as in \cref{fig:vis-gen-order} (d).
  \item Do not use anchor latents as in \cref{fig:vis-gen-order} (c).
\end{itemize}

\subsection{Inference}\label{sec:anchor-details:inference}

In \method, at inference time, we generate the anchor latents but do not include them in the final output video, like in \cref{fig:vis-gen-order} (d).
In this way, the anchor latents only serve as stabilizing current block's generation.
On the other hand, we can also include them in the final output video for visualization, i.e., directly expand block size from $B$ to $P+B$ and follow generation order as in \cref{fig:vis-gen-order} (c) at test time.
While this can trigger the inter-block flickering issue (since the anchor latents are generated without past context), it can help us visualize the anchor latents to understand what they represent.
This gives us two options in inference:
\begin{itemize}[nosep]
  \item Using anchor latents as in \cref{fig:vis-gen-order} (d).
  \item Do not use anchor latents as in \cref{fig:vis-gen-order} (c).
\end{itemize}

\subsection{Further Ablation Studies on Anchor Latents}\label{sec:anchor-details:ablation}

\begin{table}[t]
\centering
\caption{\textbf{Ablation on anchor latent configurations during training and inference.} 
We vary backward block size, anchor latent usage in Stage-1/Stage-2, loss on anchors, and whether to generate anchor latents at inference. 
Using anchor latents at inference means we generate anchor latents but do not include them in output, as in \cref{fig:vis-gen-order} (d). 
While not using anchor latents at inference means every generated latent is decoded into outputs, as in \cref{fig:vis-gen-order} (b,c).
Flickering ratio (FR) is reported for both forward ($\rightarrow$) and backward ($\leftarrow$) generation and inter/intra-block boundaries. Lower FR indicates smoother transitions.}
\label{tab:ablate-anchor-training}
\setlength{\tabcolsep}{3pt}
\resizebox{\textwidth}{!}{%
\begin{tabular}{c c c cc c c cc cc}
\toprule
\multirow{2}{*}{\#} & \multirow{2}{*}{\shortstack{Backward \\Block Size \\$B$}} & \multirow{2}{*}{\shortstack{Num Anchor \\$P$}} & \multicolumn{2}{c}{Stage-1} & \multirow{2}{*}{\shortstack{Stage-2\\Anchor}} & \multirow{2}{*}{\shortstack{Inference\\Anchor}} & \multicolumn{2}{c}{Forward FR} & \multicolumn{2}{c}{Backward FR} \\
\cmidrule(lr){4-5} \cmidrule(lr){8-9} \cmidrule(lr){10-11}
& & & Anchor & Loss on Anchor & & & Inter $\downarrow$ & Intra $\downarrow$ & Inter $\downarrow$ & Intra $\downarrow$ \\
\midrule
0 & \multicolumn{6}{c}{\textit{Self-Forcing (forward only)}} & 1.15 & 0.93 & -- & -- \\
\midrule
\multicolumn{7}{c}{\textit{Without anchor latents}} \\
1 & 3 & 0 & {\color{red}\xmark} & -- & {\color{red}\xmark} & {\color{red}\xmark} & 1.16 & 0.87 & 1.42 & 0.89 \\
2 & 6 & 0 & {\color{red}\xmark} & -- & {\color{red}\xmark} & {\color{red}\xmark} & 1.16 & 0.91 & 1.48 & 0.98 \\
3 & 6 & 0 & {\color{red}\xmark} & -- & {\color{red}\xmark} & {\color{ForestGreen}\cmark} & 1.16 & 0.91 & 1.08 & 0.98 \\
\midrule
\multicolumn{7}{c}{\textit{With anchor latents ($P{=}3$)}} \\
4 & 3 & 3 & {\color{ForestGreen}\cmark} & {\color{red}\xmark} & {\color{red}\xmark} & {\color{red}\xmark} & 1.17 & 0.90 & 1.45 & 1.04 \\
5 & 3 & 3 & {\color{ForestGreen}\cmark} & {\color{red}\xmark} & {\color{red}\xmark} & {\color{ForestGreen}\cmark} & 1.17 & 0.90 & 1.17 & 1.03 \\
6 & 3 & 3 & {\color{ForestGreen}\cmark} & {\color{red}\xmark} & {\color{ForestGreen}\cmark} & {\color{red}\xmark} & 1.13 & 0.88 & 1.46 & 0.99 \\
7 & 3 & 3 & {\color{ForestGreen}\cmark} & {\color{red}\xmark} & {\color{ForestGreen}\cmark} & {\color{ForestGreen}\cmark} & 1.13 & 0.88 & \textbf{1.07} & 0.97 \\
8 & 3 & 3 & {\color{ForestGreen}\cmark} & {\color{ForestGreen}\cmark} & {\color{red}\xmark} & {\color{red}\xmark} & 1.14 & 0.89 & 1.45 & 0.99 \\
9 & 3 & 3 & {\color{ForestGreen}\cmark} & {\color{ForestGreen}\cmark} & {\color{red}\xmark} & {\color{ForestGreen}\cmark} & 1.14 & 0.89 & 1.37 & 0.95 \\
8 & 3 & 3 & {\color{ForestGreen}\cmark} & {\color{ForestGreen}\cmark} & {\color{ForestGreen}\cmark} & {\color{red}\xmark} & 1.13 & 0.91 & 1.38 & 0.99 \\
9 & 3 & 3 & {\color{ForestGreen}\cmark} & {\color{ForestGreen}\cmark} & {\color{ForestGreen}\cmark} & {\color{ForestGreen}\cmark} & 1.13 & 0.91 & 1.15 & 0.95 \\
\bottomrule
\end{tabular}%
}
\end{table}

\begin{table}[t]
\centering
\caption{\textbf{Ablation on the number of anchor latents $P$.} All models follow the same configuration as EXP-7 in \cref{tab:ablate-anchor-training}, except for the number of anchor latents $P$. 
Increasing $P$ progressively reduces backward inter-block FR, with $P{=}3{=}B$ achieving the best result.}
\label{tab:ablate-anchor-num}
\setlength{\tabcolsep}{5pt}
\begin{tabular}{c c cc cc}
\toprule
\multirow{2}{*}{\#} &\multirow{2}{*}{\shortstack{Num Anchor\\$P$}} & \multicolumn{2}{c}{Forward FR} & \multicolumn{2}{c}{Backward FR} \\
\cmidrule(lr){3-4} \cmidrule(lr){5-6}
& & Inter $\downarrow$ & Intra $\downarrow$ & Inter $\downarrow$ & Intra $\downarrow$ \\
\midrule
10 & 1 & 1.13 & 0.90 & 1.26 & 0.93 \\
11 & 2 & 1.09 & 0.91 & 1.23 & 0.93 \\
7 & 3 & 1.13 & 0.88 & \textbf{1.07} & 0.97 \\
\bottomrule
\end{tabular}
\end{table}

We provide additional ablation studies on anchor latents beyond those in the main paper, by training/testing models after applying different design choices in stage-1 training, stage-2 training, and inference as discussed above.
We provide quantitative results in \cref{tab:ablate-anchor-training} and \cref{tab:ablate-anchor-num}.
Besides, we also provide visualizations in the attached main.html file.

\medskip
\noindent\textbf{Including anchor latents as outputs at inference time.}
As shown in \cref{tab:ablate-anchor-training}, we find it is crucial to discard the generated anchor latents to allow different blocks overlap during inference, no matter how the model is trained.
For the baseline that does not incorporate anchor latents during training and uses block size $B=6$ in backward generation,
  we can still discard the first 3 latents generated in each block, simulating the anchor latent effect at test time by following the generation order in \cref{fig:vis-gen-order}~(d).
This simple step can largely help alleviate the inter-block flickering issue, decreasing the backward inter-block FR from $1.48$ to $1.08$.
This also gives another interpretation of the anchor latents: they enable overlaping different blocks' generation, thus improving continuity of the generated video.

\medskip
\noindent\textbf{Use of anchor latents during training.}
As shown in \cref{tab:ablate-anchor-training}, it is important to keep stage-1 and stage-2 consistent.
When we introduce anchor latents in one stage but not the other, the performance will degrade (e.g., from $1.07$ in Exp-7 to $1.17$ in Exp-5 and $1.37$ in Exp-9 for backward generation).
Also, adding loss to anchor latents in stage-1 can slightly hurt performance (from $1.07$ in Exp-7 to $1.15$ in Exp-9 for backward generation).
We hypothesize that this is because, at test time, the anchor latents are not decoded but only help in stailizing current block's generation. Therefore, adding loss on them is uncecessary and can distract the model from optimizing current block's generation.

\medskip
\noindent\textbf{Effects of anchor latent numbers $P$.}
\cref{tab:ablate-anchor-num} ablates the number of anchor latents $P \in \{1, 2, 3\}$.
Increasing $P$ progressively reduces backward inter-block FR: $P{=}1$ ($1.26$), $P{=}2$ ($1.23$), $P{=}3$ ($\mathbf{1.07}$).
At $P{=}3{=}B$, the anchor provides a full block of past context, and the ratio approaches the ideal value of $1.0$.

\medskip
\noindent\textbf{What do anchor latents represent?}
By decoding anchor latents into actual videos following the generation order in \cref{fig:vis-gen-order} (c), as in Exp-6 of \cref{tab:ablate-anchor-training} and the attached main.html file, we can observe that the anchor latents can be decoded into video latents.
This shows they are indeed approximation of the past video latents.
This further verifies our motivation that the generation of each latent should be conditioned on some past context.

Our final model corresponds to Exp-7 in \cref{tab:ablate-anchor-training}, where we introduce anchor latents in Stage-1 and Stage-2 and discard them at inference, without applying loss on anchor latents in Stage-1.

\section{Training and Inference Efficiency}\label{sec:appendix-efficiency}

We provide detailed training and inference efficiency measurements in \cref{tab:suppl-efficiency}.
\definecolor{lightgreen}{RGB}{232,245,233}

\begingroup
\centering
\renewcommand{\arraystretch}{0.92}
\captionsetup{type=table,hypcap=false}
\caption{\textbf{Training and inference efficiency.} We report training cost, autoregressive video generation inference cost, and inbetween generation inference cost. Training is measured per GPU on 8x8 H100s; inference is measured on a H100.}
\label{tab:suppl-efficiency}

\begin{minipage}[t]{\textwidth}
\centering
\setlength{\tabcolsep}{5pt}
\begin{tabular}{lccccc}
\toprule
\shortstack{Model\\\vphantom{Latency(s)}} & \shortstack{Past\\Sink} & \shortstack{Future\\Sink} & \shortstack{FPS$\uparrow$\\\vphantom{Latency(s)}} & \shortstack{First-frame$\downarrow$\\Latency(s)} & \shortstack{VRAM$\downarrow$\\(GiB)} \\
\midrule
LongLive~\cite{longlive2025} & \checkmark & $\times$ & 12.6 & 0.79 & 20.39 \\
RollingForcing(RF)~\cite{rollingforcing2025} & \checkmark & $\times$ & 4.5 & 5.58 & 22.99 \\
Infinity-Rope~\cite{infinityrope2025} & \checkmark & $\times$ & 16.2 & 0.61 & 17.37 \\
Self-Forcing~\cite{selfforcing} & \checkmark & $\times$ & 14.8 & 0.61 & 19.18 \\
\rowcolor{lightgreen}\method-F & \checkmark & $\times$ & 14.8 & 0.61 & 19.18 \\
\rowcolor{lightgreen}\method-F & \checkmark & \checkmark & 13.6 & 0.61 & 20.14 \\
\rowcolor{lightgreen}\method-B & $\times$ & \checkmark & 9.3 & 0.71 & 20.16 \\
\rowcolor{lightgreen}\method-B & \checkmark & \checkmark & 8.6 & 0.71 & 21.11 \\
VACE~\cite{vace2025} & $\times$ & $\times$ & 0.08 & 157.14 & 27.98 \\
VACE~\cite{vace2025} & $\times$ & $\times$ & 0.07 & 160.84 & 27.98 \\
\bottomrule
\end{tabular}%
\par
\vspace{0.2em}
\footnotesize\textbf{(a) Autoregressive video generation inference cost.}
\end{minipage}\hfill

\vspace{0.25em}
\begin{minipage}[t]{0.48\textwidth}
\centering
\setlength{\tabcolsep}{5pt}
\resizebox{\linewidth}{!}{%
\begin{tabular}{cccc}
\toprule
Generator & Unified & VRAM(GiB) & Time(hr) \\
$G^\theta$ & $\mu_\text{fake}^\phi$ & Stage-1 / 2 & Stage-1 / 2 \\
\midrule
Forward only & -- & 47.3 / 63.5 & 11 / 1.5 \\
Backward only & -- & 47.3 / 72.0 & 11 / 2.0 \\
Unified & $\times$ & 51.9 / 76.3 & 16 / 3.5 \\
\rowcolor{lightgreen}Unified & \checkmark & 51.9 / 72.0 & 16 / 3.0 \\
\bottomrule
\end{tabular}%
}
\vspace{0.2em}
\footnotesize\textbf{(b) Per-GPU training cost on 8x8 H100s.}
\end{minipage}\hfill
\begin{minipage}[t]{0.49\textwidth}
\centering
\setlength{\tabcolsep}{4pt}
\resizebox{\linewidth}{!}{%
\begin{tabular}{lcc}
\toprule
Model & Params(B) & Time(s) \\
\midrule
Wan FLF2V~\cite{wan2024} & 14 & 1966.3 \\
GI~\cite{gi2024} & 1.5 & 295.7 \\
\rowcolor{lightgreen}\method & 1.3 & 9.0 \\
\bottomrule
\end{tabular}%
}
\vspace{0.2em}
\footnotesize\textbf{(c) Inbetween generation inference cost.}
\end{minipage}
\par
\endgroup

\section{Detailed Results}\label{sec:appendix-results}

In this section, we include more quantitative evaluation results on short video generation in \cref{tab:short-video-gen-full}.

\begin{table}[!htbp]
\centering
\caption{\textbf{Detailed VBench scores for single-direction short video generation.} We report all 16 VBench dimensions along with the aggregated quality, semantics, and total scores. 
 Our unified training framework helps obtain a model for both forward and backward generation while maintaining similar performance compared to single-direction only models.}
\label{tab:short-video-gen-full}
\setlength{\tabcolsep}{3pt}
\resizebox{\textwidth}{!}{%
\begin{tabular}{l cccccccccccccccc|ccc}
\toprule
Model
& \rotatebox{70}{subject consistency} & \rotatebox{70}{background consistency} & \rotatebox{70}{aesthetic quality} & \rotatebox{70}{image quality}
& \rotatebox{70}{object class} & \rotatebox{70}{multi-object} & \rotatebox{70}{color} & \rotatebox{70}{spatial relation}
& \rotatebox{70}{scene} & \rotatebox{70}{temporal style} & \rotatebox{70}{overall consistency} & \rotatebox{70}{human action}
& \rotatebox{70}{temporal flicker} & \rotatebox{70}{motion smoothness} & \rotatebox{70}{dynamic} & \rotatebox{70}{appearance style}
& \rotatebox{70}{quality} & \rotatebox{70}{semantics} & \rotatebox{70}{total} \\
\midrule
Forward only (Self-Forcing)       & 96.54 & 95.83 & 66.10 & 69.81 & 93.64 & 80.87 & 87.39 & 76.78 & 54.45 & 23.87 & 26.60 & 96.60 & 98.26 & 98.31 & 65.00 & 20.86 & 84.47 & 79.23 & 83.42 \\
\method - Forward  & 95.94 & 95.15 & 65.73 & 69.34 & 93.62 & 82.21 & 86.71 & 77.09 & 56.45 & 24.01 & 26.52 & 96.80 & 97.67 & 97.85 & 81.67 & 20.49 & 84.89 & 79.51 & 83.81 \\
\method - Backward & 96.67 & 95.95 & 66.90 & 70.14 & 95.21 & 86.20 & 87.19 & 81.49 & 56.42 & 24.15 & 26.83 & 95.20 & 98.51 & 98.10 & 70.56 & 20.66 & 85.12 & 80.69 & 84.23 \\
Backward only      & 96.13 & 95.45 & 65.89 & 69.54 & 95.32 & 86.91 & 88.88 & 78.49 & 56.58 & 24.02 & 26.54 & 95.60 & 98.13 & 97.65 & 82.22 & 20.74 & 85.17 & 80.61 & 84.26 \\
\bottomrule
\end{tabular}%
}
\end{table}

\FloatBarrier
\section{Subjective Assessment}\label{sec:appendix-subjective}

We conduct a user study with A/B tests to subjectively assess cross-block flickering.
Each user watches 20 random pairs of videos generated with and without anchor latents, then chooses the one with less cross-block flickering.
\cref{tab:suppl-user-study} shows an 87\% preference for videos generated with anchor latents and 80.5\% agreement with our flickering ratio (FR).

\begin{table}[!htbp]
\centering
\caption{\textbf{User study for evaluating cross-block flickering.} We report preference between videos generated with and without anchor latents, together with agreement between user preference and the proposed flickering ratio (FR).}
\label{tab:suppl-user-study}
\setlength{\tabcolsep}{3.5pt}
\resizebox{\textwidth}{!}{%
\begin{tabular}{cccc}
\toprule
\raisebox{-0.55\normalbaselineskip}[0pt][0pt]{Number of Responses} & \multicolumn{2}{c}{Preference Rate} & \raisebox{-0.55\normalbaselineskip}[0pt][0pt]{Agreement with FR} \\
\cmidrule(lr){2-3}
& w/ Anchor & w/o Anchor & \\
\midrule
200 & 87\% & 13\% & 80.5\% \\
\bottomrule
\end{tabular}%
}
\end{table}

\FloatBarrier

\end{document}